\PassOptionsToPackage{numbers,sort&compress}{natbib}
\documentclass{article}

\pdfoutput=1

\usepackage[preprint]{corl_2025}
\usepackage{times}
\usepackage{wrapfig}
\usepackage{pdfx}

\usepackage[dvipsnames,table,xcdraw, HTML]{xcolor}

\usepackage{hyperref}
\hypersetup{
    colorlinks=true,
    linkcolor=orange,
    filecolor=orange,      
    urlcolor=orange,
    citecolor=orange,
}

\usepackage{url}
\usepackage{booktabs}
\usepackage{graphicx}
\usepackage{array}
\usepackage{multirow}
\usepackage{wrapfig}
\usepackage{amsmath}
\usepackage{amsfonts}
\usepackage{amssymb}
\usepackage[capitalise, nameinlink]{cleveref}
\usepackage{comment}
\usepackage{inconsolata}   
\usepackage{xcolor} 

\definecolor{darkgreen}{RGB}{0,100,0} 
\definecolor{lightgreen}{RGB}{144,238,144} 

\usepackage[tableposition=top]{caption}
\usepackage{multicol}
\usepackage{subcaption}
\usepackage{environ}
\usepackage{tabularx}
\usepackage{authblk}
\usepackage{xspace}
\usepackage{bbding}
\usepackage{pgfplots}
 \usepackage{pifont} 
\pgfplotsset{compat=1.18} 
\usepackage{pgfplotstable}
\usepackage{xcolor}
\usepackage{tikz}
\usetikzlibrary{patterns}

\usepackage{titlesec}
\titlespacing\section{0pt}{0pt plus 2pt minus 2pt}{0pt plus 2pt minus 2pt}
\titlespacing\subsection{0pt}{3pt plus 4pt minus 2pt}{0pt plus 2pt minus 2pt}
\titlespacing\subsubsection{0pt}{3pt plus 4pt minus 2pt}{0pt plus 2pt minus 2pt}

\usepackage[labelfont=bf,labelsep=period]{caption}

\title{\acro: Learning In-the-Wild Mobile Manipulation via Hybrid Imitation and Whole-Body Control} 

\author{
\textbf{Priya Sundaresan$^1$}, 
\textbf{Rhea Malhotra$^1$},
\textbf{Phillip Miao$^1$},
\textbf{Jingyun Yang$^1$}, 
\textbf{Jimmy Wu$^1$}, \\
\textbf{Hengyuan Hu$^1$}, 
\textbf{Rika Antonova$^{1,2}$},  
\textbf{Francis Engelmann$^1$},  
\textbf{Dorsa Sadigh$^1$}, 
\textbf{Jeannette Bohg$^1$}
}

\affil{$^1$Stanford University, $^2$University of Cambridge}

\makeatletter
\def\thanks#1{\protected@xdef\@thanks{\@thanks
        \protect\footnotetext{#1}}}
\makeatother

\begin{document}

\newcommand\smallO{
  \mathchoice
    {{\scriptstyle\mathcal{O}}}
    {{\scriptstyle\mathcal{O}}}
    {{\scriptscriptstyle\mathcal{O}}}
    {\scalebox{.6}{$\scriptscriptstyle\mathcal{O}$}}
  }

\def\colorModel{hsb} 

\newcommand\ColCell[1]{
  \pgfmathparse{#1<50?1:0}  
    \ifnum\pgfmathresult=0\relax\color{white}\fi
  \pgfmathsetmacro\compA{0}      
  \pgfmathsetmacro\compB{#1/100} 
  \pgfmathsetmacro\compC{1}      
  \edef\x{\noexpand\centering\noexpand\cellcolor[\colorModel]{\compA,\compB,\compC}}\x #1
  } 
\newcolumntype{E}{>{\collectcell\ColCell}m{0.4cm}<{\endcollectcell}}  
\newcommand*\rot{\rotatebox{90}}

\newcommand{\Exp}{\mathbb{E}}
\newcommand{\inlineeqnum}{\refstepcounter{equation}~~\mbox{(\theequation)}}
\newcommand{\eg}{\emph{e.g.}\xspace}
\newcommand{\ie}{\emph{i.e.}\xspace}
\newcommand{\acro}{\textsc{HoMeR}}

\captionsetup[figure]{name=Fig.}
\captionsetup[table]{name=Tab.}

\maketitle

\vspace{-0.75cm}

\begin{abstract}
We introduce {\acro}, an imitation learning framework for mobile manipulation that combines whole-body control with hybrid action modes that handle both long-range and fine-grained motion, enabling effective performance on realistic in-the-wild tasks. At its core is a fast, kinematics-based whole-body controller that maps desired end-effector poses to coordinated motion across the mobile base and arm. Within this reduced end-effector action space, \acro{} learns to switch between absolute pose predictions for long-range movement and relative pose predictions for fine-grained manipulation, offloading low-level coordination to the controller and focusing learning on task-level decisions.
We deploy \acro{} on a holonomic mobile manipulator with a 7-DoF arm in a real home. We compare \acro{} to baselines without hybrid actions or whole-body control across 3 simulated and 3 real household tasks such as opening cabinets, sweeping trash, and rearranging pillows. Across tasks, \acro{} achieves an overall success rate of $\mathbf{79.17\%}$ using just 20 demonstrations per task, outperforming the next best baseline by $\mathbf{29.17\%}$ on average. \acro~ is also compatible with vision-language models and can leverage their internet-scale priors to better generalize to novel object appearances, layouts, and cluttered scenes. In summary, \acro{} moves beyond tabletop settings and demonstrates a scalable path toward sample-efficient, generalizable manipulation in everyday indoor spaces. Code, videos, and supplementary material are available at: \href{http://homer-manip.github.io}{\texttt{http://homer-manip.github.io}}.
\end{abstract}


\begin{figure*}[!htbp]
    \centering
    \vspace{-0.3cm}
    \includegraphics[width=1.0\linewidth]{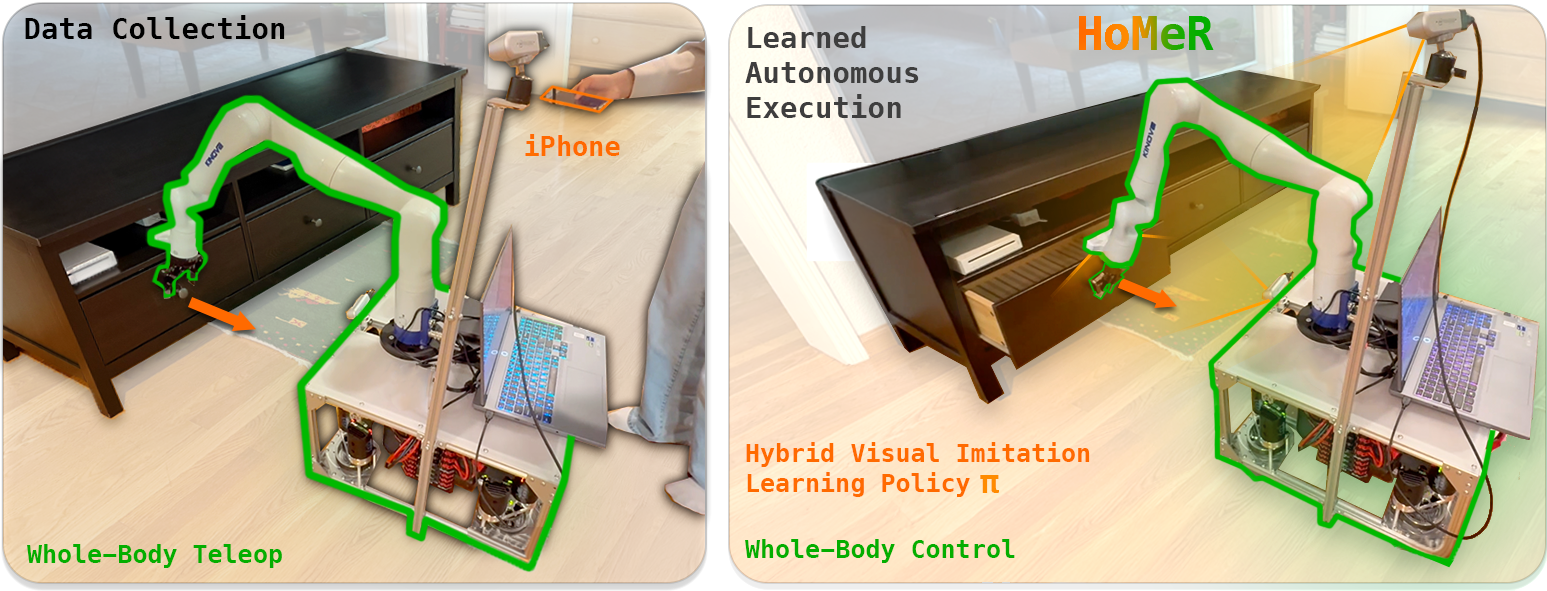}
    \caption{\small{\textbf{\acro.} 
Left: A demonstrator uses whole-body iPhone teleoperation to collect data with a mobile manipulator in a real home.
Right: From these collected demonstrations, \acro{} learns a hybrid imitation learning policy that switches between absolute actions for reaching, and relative actions for fine manipulation. A whole-body controller maps these end-effector commands to arm and base joint commands for execution.}}
    \label{fig:splash}
    \vspace{-0.1cm}
\end{figure*}
\section{Introduction}
\label{sec:intro}

To unlock the full potential of robots, we must move beyond controlled lab spaces and into the diverse, unstructured environments of everyday life.
Unlike stationary tabletop robots, which lack the mobility to perform the wide range of tasks found in homes, offices, and warehouses, mobile manipulators are capable of navigating and interacting in these human-centric spaces.
Tasks like watering a row of plants, wiping a spill on a long table, or moving to open a cabinet (\cref{fig:splash}) require not only precision, but also manipulation and mobility working hand-in-hand.

Modern mobile embodiments such as wheeled base-arm platforms~\cite{wu2023tidybot, wu2024tidybot++, brondmo2019everyday, haviland2022holistic, nasiriany2024robocasa, rana2023sayplan, wise2016fetch, yamamoto2018human}, humanoids~\cite{bjorck2025gr00t, cheng2024expressive, qiu2025humanoid, fu2024humanplus, 1x2025worldmodel, wang2025mobile, wang2024exbody2, unitree2025h1, sferrazza2024humanoidbench}, and quadrupeds~\cite{shah2022lmnav, lee2020learning, kumar2021rma, agarwal2023legged, margolis2022learning} enable robots to operate beyond fixed workspaces. However, these embodiments introduce significant control complexity. In particular, they require careful coordination between the base and arm in wheeled embodiments, or the limbs and torso in legged systems. One common approach to managing this complexity is to use whole-body controllers (WBCs), which map high-level end-effector commands to coordinated whole-body motion using analytical or learning-based approaches. Prior work on WBCs has focused primarily on quadrupeds~\cite{ha2024umi, fu2022deep, liu2024visual, bruedigam2024jacta}, where achieving balance and stability requires learning WBCs from scratch with extensive reward shaping and sim-to-real transfer techniques, often tailored to the dynamics of legged robots.
Furthermore, real-world tasks are naturally multi-phase, combining long-range movements (e.g., reaching or repositioning) with fine-grained local manipulation of objects.

On the policy learning side, recent imitation learning (IL) methods~\cite{sundaresan2024s,shi2023waypoint, belkhale2023hydra} have shown the benefits of hybrid action spaces, although their application has been restricted to tabletop domains. These policies typically learn to alternate between predicting (1) \emph{keypose} actions (6 DoF absolute end-effector poses, equivalently referred to as \emph{waypoint}~\cite{sundaresan2024s, shi2023waypoint, belkhale2023hydra} or \emph{keyframe}~\cite{shridhar2023perceiver, james2022q} actions) for long-range movement and (2) \emph{dense} actions (\ie, delta end-effector actions) for fine-grained interactions. While these approaches achieve strong performance and generalization from limited demonstrations in tabletop settings, they have not been introduced in the mobile manipulation space with greater embodiment complexity, larger workspaces, and longer task horizons.

Our key insight is that scalable mobile manipulation requires not only an effective strategy for managing control complexity, but also a means of generalizing to novel scenarios. As mobile robots move through diverse human environments, they are exposed to far greater variability in objects, spatial configurations, and task conditions than static arms. To address both challenges, we propose \textbf{\acro}: \underline{H}ybrid wh\underline{o}le-body policies for \underline{M}obil\underline{e} \underline{R}obots. \acro{} (\cref{fig:splash}) combines a fast, kinematics-based whole-body controller (which maps end-effector actions to coordinated base-arm motion) with a hybrid IL policy that switches between keypose predictions for long-range movement and dense delta actions for fine-grained manipulation. This structured approach addresses high-dimensional control and multi-phase execution. Additionally, we show that \acro{} is modular enough to incorporate task-relevant keypoints derived from vision-language models (VLMs), providing a path towards improved generalization in unfamiliar environments.

We deploy \textbf{\acro} in a real home environment and evaluate it on a suite of challenging mobile manipulation tasks that reflect everyday household demands. 
Overall, our contributions are:
\begin{enumerate}
    \item A \textbf{sample-efficient imitation learning framework for mobile manipulation} that leverages WBC and hybrid action representations to outperform strong non-hybrid and non-WBC baselines using only 20 demonstrations per task.
    \item A \textbf{modular policy architecture that can be conditioned on VLM keypoints}, enabling generalization to novel object geometries, appearances, and cluttered environments.
    \item A \textbf{practical whole-body controller that supports intuitive teleoperation}, facilitating efficient demonstration collection in real household settings.
\end{enumerate}

\section{Related Work}
\label{sec:related}

\paragraph{Mobile Manipulation and Control.}  
\emph{Legged platforms} typically focus on navigation across diverse terrains (\eg, with Boston Dynamics Spot~\cite{agarwal2023legged} and ANYmal~\cite{hutter2017anymal} quadrupeds). A few recent works equip quadrupeds with lightweight arms and develop whole-body controllers, utilizing either model-based motion planning (\eg, RoLoMa~\cite{ferrolho2023roloma}) or learning-based manipulation policies (\eg, DeepWBC~\cite{fu2022deep}, Visual WBC~\cite{liu2024visual}, UMI-on-Legs~\cite{ha2024umi}). 
Our work draws inspiration from these works, but our framework is agnostic to the exact implementation of the WBC (\eg, inverse kinematics (IK) or learning-based). In our work, we use a task-agnostic IK-based WBC that is reusable across many scenarios. Furthermore, in contrast to these prior works, we learn policies with hybrid action modes to encourage both spatial generalization and precision.
More recently, \emph{humanoid} research has advanced rapidly, with most efforts centered on whole-body control for expressive behaviors such as dancing, walking, or jumping~\cite{cheng2024expressive, qiu2025humanoid}.
Although some humanoid systems perform manipulation, they often rely on using high-dimensional joint-space actions (\eg, 50+ DoF for HumanPlus~\cite{fu2024humanplus}) for imitation learning. In contrast, our work adopts a whole-body control strategy with hybrid action modes to enable more tractable learning.

\emph{Wheeled platforms} consist of a wheeled mobile base with onboard arms, and include examples like  TidyBot~\cite{wu2023tidybot}, TidyBot++~\cite{wu2024tidybot++}, mobile Franka Pandas~\cite{haviland2022holistic, nasiriany2024robocasa, rana2023sayplan}, the Fetch robot~\cite{wise2016fetch}, the HSR~\cite{yamamoto2018human}, Mobile Aloha~\cite{fu2024mobile}, and the Everyday Robot~\cite{brohan2022rt}. Although these embodiments are physically capable of performing many mobile manipulation tasks, they typically assume decoupled control of the base and arm.
Having to switch between different movement modes adds complexity to teleoperation, and often requires the use of ad hoc and task-specific strategies.

A variety of other works study \emph{navigation} with legged or wheeled embodiments~\cite{hirose2024lelan, yokoyama2024vlfm, jauhri2024active}. Our work is different and complementary in that we focus on the ``last mile'' of manipulation, where mobility is necessary for task completion, but not at the scale or complexity of full-scene navigation.

\paragraph{Visual Imitation Learning.} Visual imitation learning (IL) refers to learning from demonstrations using visual observations~\cite{schaal1996learning, ross2011reduction, levine2016end}, with 
recent methods exploring various alternatives for action granularity and input/output space structure.

\emph{Dense policies}, such as Diffusion Policy~\cite{chi2023diffusion}, ACT~\cite{zhao2023learning}, or Visual-Language-Action (VLA) models (Gemini~\cite{team2025gemini}, $\pi_{0}$~\cite{black2024pi_0}, RT-X~\cite{open_x_embodiment_rt_x_2023}, OpenVLA~\cite{kim24openvla}, Octo~\cite{team2024octo}) predict low-level actions (\eg{}, 6-DoF deltas or joint velocities) at every timestep. While effective for reactive manipulation, dense policies struggle with long-horizon tasks and spatial generalization, since even simple movements like reaching can involve hundreds of consecutive actions.

\emph{Keypose-based policies}, such as PerAct~\cite{shridhar2023perceiver}, RVT/RVT-2~\cite{goyal2023rvt,goyal2024rvt}, and KITE~\cite{sundaresan2023kite}, predict 6-DoF end-effector poses that are executed via low-level controllers or motion planners. While sample-efficient, these keypose actions can be too sparse to handle precise or reactive control.

Recently proposed \emph{hybrid policies} combine keypose and dense actions for both long-range and precise local manipulation.
Hydra~\cite{belkhale2023hydra} and AWE~\cite{shi2023waypoint} use keypose and/or dense actions but rely solely on images, limiting spatial generalization.
SPHINX~\cite{sundaresan2024s}, most similar to our approach, uses images and point clouds with learned attention to task-relevant keypoints to switch between modes.
Critically, all these methods are limited to static, tabletop-mounted manipulators.
We extend SPHINX to the mobile manipulation setting by incorporating whole-body control, enabling mobility while retaining an end-effector–centric action space. We further support optional conditioning on object keypoints from VLMs, allowing generalization to unseen objects in clutter.

\begin{figure*}[t]
    \centering
    \includegraphics[width=1.0\linewidth]{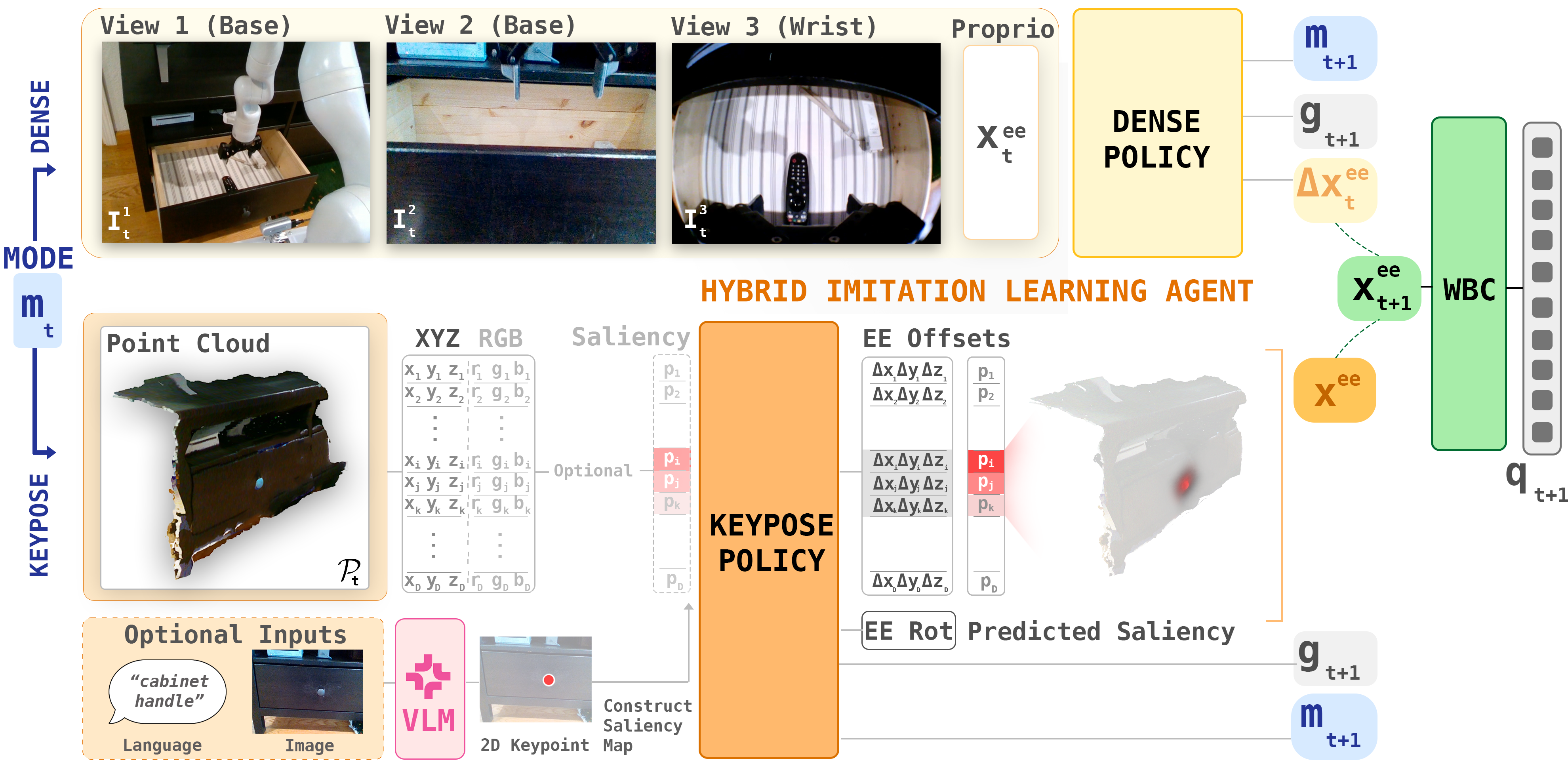}
    \caption{\small{\textbf{\acro~Policy Architecture:} \acro{} consists of a \emph{dense policy} that uses RGB images to predict relative actions for fine-grained manipulation, and a \emph{keypose policy} that uses point clouds to predict absolute end-effector poses for long-range motion. Each policy also predicts the next control mode, enabling learned transitions. Optionally, the keypose policy can be conditioned on externally provided salient points derived from a VLM to support dynamic goal specification (\acro-\textsc{Cond}). Finally, a whole-body controller (WBC) converts predicted end-effector actions into joint commands for the mobile base and arm.}}
    \label{fig:method}
    \vspace{-1em}
\end{figure*}

\vspace{5px}
\section{\acro: \sc{\underline{H}ybrid wh\underline{O}le-body policies for \underline{M}obil\underline{E} \underline{R}obots}}
\label{sec:method}

In the following sections, we describe \acro{} (\cref{fig:method}), our imitation learning framework for mobile manipulation in the wild.
\cref{sec:method-setup} formalizes the problem setup.
\cref{sec:method-wbc} presents a kinematics-based \emph{whole-body controller (WBC)} that enables control in a simplified end-effector action space.
Finally, \cref{sec:method-policy} describes our \emph{hybrid imitation learning (IL) agent}, which maps point clouds and RGB image inputs to hybrid actions for both long-range and fine-grained manipulation.

\subsection{Problem Formulation}
\label{sec:method-setup}

We consider a mobile manipulator composed of a holonomic mobile base and an $N$-DoF robotic arm, with joint configuration $
\mathbf{q}_t = (\mathbf{q}^{\textbf{base}}_t, \mathbf{q}^{\textbf{arm}}_t) \in \mathbb{R}^{3+N},
$
where
\(\mathbf{q}^{\textbf{base}}_t = (x, y, \theta) \in SE(2)\)
represents the base pose, and \(\mathbf{q}^{\textbf{arm}}_t \in \mathbb{R}^N\) represents the arm joints. 

At each timestep \(t\), an observation
$
o_t \!=\! \left(\mathbf{q}_t, g_t, \{\mathbf{I}_t^{k}, \mathbf{D}_t^{k}\}_{k=1}^K \right)
$
includes the joint configuration $\mathbf{q}_t$, gripper state \(g_t \!\in\! \mathbb{R}\), and RGB-D images from \(K\) cameras (with at least one wrist-mounted and one third-person view). To get 3D point clouds, we assume known camera intrinsics and extrinsics.

Rather than learning actions directly in joint space, our goal is to train an imitation learning (IL) policy \(\pi(o_t) = (\mathbf{x}^{\text{ee}}_{t+1}, g_{t+1})\) that predicts a 6-DoF end-effector target pose \(\mathbf{x}^{\text{ee}}_{t+1} \!\in\! SE(3)\), which can then be executed by a whole-body controller (WBC). With this formulation, the IL policy reasons in task space while delegating low-level control and embodiment-specific constraints to the WBC.

\subsection{Whole-Body Controller}
\label{sec:method-wbc}

We implement a kinematics-based WBC that maps high-level end-effector poses into joint position commands for the full embodiment, delegating joint-space coordination, constraints, and redundancy resolution to the controller rather than the IL agent. Though \acro~is agnostic to the exact WBC implementation, ours is based on MuJoCo~\cite{todorov2012mujoco} and the mink IK library~\cite{mink}.

Formally, the WBC is a mapping \(\mathcal{W} \!:\! SE(3) \rightarrow \mathbb{R}^{3+N}\) from a desired end-effector pose \(\mathbf{x}^{\text{ee}} \!\in\! SE(3)\) to a joint position command \(\mathbf{q} \in \mathbb{R}^{3+N}\) for the mobile base and arm. 
We implement an IK solver based on iterative IK that finds $\mathbf{q}$ minimizing the pose error.
To compute a velocity that moves the end-effector toward \(\mathbf{x}^{\text{ee}}\), we define a pose error as a body-frame twist~\cite{murray1994mathematical}:
\[
\mathbf{e}^{\text{ee}} = \log\left((\mathbf{x}_t^{\text{ee}})^{-1} \mathbf{x}^{\text{ee}}\right),
\]
where \(\mathbf{x}_t^{\text{ee}} \in SE(3)\) is the current end-effector pose from forward kinematics. The geometric Jacobian \(\mathbf{J}_{\text{ee}}(\mathbf{q}_t) \in \mathbb{R}^{6 \times (3+N)}\) maps joint velocities to the induced end-effector twist. At each iteration, the IK solver finds \(\dot{\mathbf{q}}\) that minimizes the discrepancy between the Jacobian-induced twist and \(\mathbf{e}^{\text{ee}}\), moving the end-effector toward the desired pose. Specifically, the IK solver optimizes the following:
\[
\begin{aligned}
\min_{\dot{\mathbf{q}}} \quad &
\left\| \mathbf{J}_{\text{ee}}(\mathbf{q}_t) \dot{\mathbf{q}} - \mathbf{e}^{\textbf{ee}} \right\|^2_{W_{\text{ee}}}
+ \left\| \mathbf{q}_t + \dot{\mathbf{q}} \cdot \Delta t - \mathbf{q}_{\text{retract}} \right\|^2_{W_{\text{posture}}}
+ \left\| \dot{\mathbf{q}}^{\textbf{base}} \right\|^2_{W_{\text{damping}}}
\end{aligned}
\]

\begin{subequations}
\begin{align}
\text{s.t.} \quad 
& \dot{\mathbf{q}}_{\min} \leq \dot{\mathbf{q}} \leq \dot{\mathbf{q}}_{\max} && \text{(a) Joint velocity limits} \label{eq:vel_limit} \\
& \mathbf{q}_{\min} \leq \mathbf{q}_t + \dot{\mathbf{q}} \cdot \Delta t \leq \mathbf{q}_{\max} && \text{(b) Joint position limits} \label{eq:pos_limit} \\
& -\mathbf{n}_i^\top \mathbf{J}_i(\mathbf{q}_t) \dot{\mathbf{q}} \leq 
\underbrace{\frac{\gamma (d_i - d_{\min})}{\Delta t} + \epsilon}_{\text{collision margin}}, 
\quad \forall i \in \mathcal{C} && \text{(c) Collision avoidance} \label{eq:collision_limit}
\end{align}
\end{subequations}

\begin{wrapfigure}{r}{0.42\textwidth}
  \centering
  \vspace{-8pt}
\includegraphics[width=0.42\textwidth,trim=0pt 0pt 0pt 0pt]{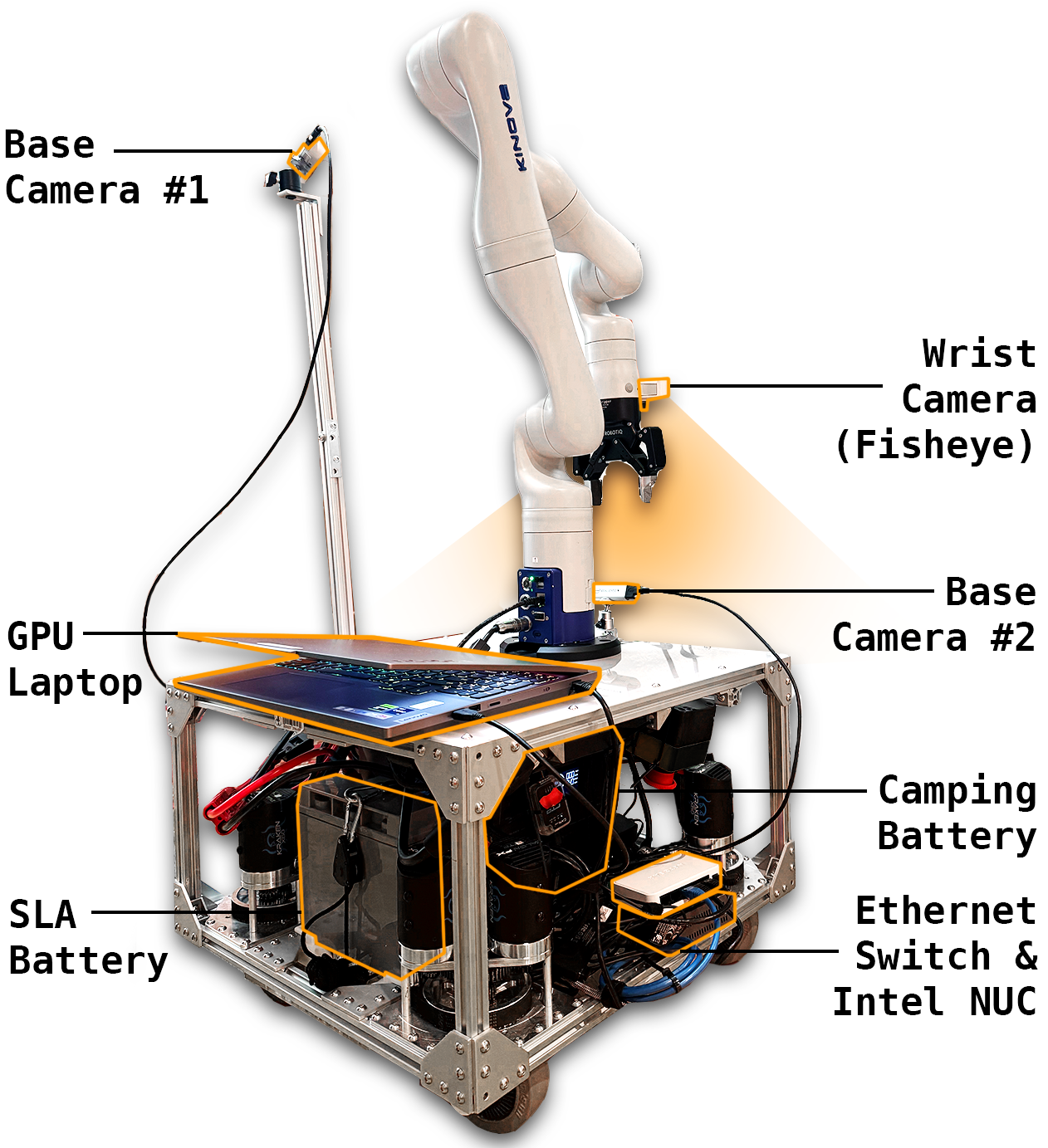}
  \caption{\small{\textbf{Hardware:} We use the TidyBot++ holonomic mobile manipulator~\cite{wu2024tidybot++} with two base cameras and a wrist-mounted fisheye camera. An onboard NUC handles real-time control, and an onboard GPU laptop runs policy inference.}}
  \vspace{-3pt}
\label{fig:hardware_setup}
\end{wrapfigure}

\textbf{Objective:} The first term encourages the solved joint motion to move towards the target pose \(\mathbf{x}^{\textbf{ee}}\). The second term encourages the joint configuration \(\mathbf{q}_t + \dot{\mathbf{q}} \cdot \Delta t\) to stay close to a neutral resting posture \(\mathbf{q}_{\text{retract}} \in \mathbb{R}^{3+N}\), shown in \cref{fig:hardware_setup}. The third term damps the motion of the base. Weights \(W_{\text{ee}}\), \(W_{\text{posture}}\), and \(W_{\text{damping}}\) specify the influence of each term.

\setlength{\intextsep}{0pt}
\setlength{\columnsep}{10pt}

\textbf{Constraints:}
The optimization is subject to constraints that ensure safe and feasible execution.
We impose joint velocity \cref{eq:vel_limit} and position limits \cref{eq:pos_limit}, \(\dot{\mathbf{q}}_{\min} \leq \dot{\mathbf{q}} \leq \dot{\mathbf{q}}_{\max}\) and \(\mathbf{q}_{\min} \leq \mathbf{q}_t + \dot{\mathbf{q}} \cdot \Delta t \leq \mathbf{q}_{\max}\), to satisfy hardware bounds.
In constraint~\cref{eq:collision_limit}, we enforce velocity-based collision avoidance between selected pairs of robot components, each modeled as geometric primitives (\emph{geoms}) in the MuJoCo simulator~\cite{todorov2012mujoco}.
For each pair \(i \in \mathcal{C}\), we identify the closest points between geoms and compute the signed distance \(d_i\), the contact normal \(\mathbf{n}_i\), and the Jacobian \(\mathbf{J}_i\) of the contact point with respect to joint motion.
The constraint aims to slow the robot's motion as the clearance \(d_i\) between geoms approaches a minimum threshold, effectively acting as a velocity damper.
In our setup, \(\mathcal{C}\) includes the (arm, base) and (arm, camera mount) pairs, where the camera mounts are represented as cylinders.

The IK solver iteratively integrates the optimized joint velocities using a fixed timestep \(\Delta t\) to obtain the final joint position command: $\mathbf{q}_{t+1} = \mathbf{q}_t + \dot{\mathbf{q}} \cdot \Delta t$.
The joint position commands are subsequently executed on hardware using low-level controllers.
All WBC hyperparameters are given in \cref{sec:appendix-wbc}. These values are held constant and reused across all tasks without any per-task retuning.

\subsection{Hybrid Imitation Learning Agent} \label{sec:method-policy}
Built on the WBC’s end-effector control space, our hybrid imitation learning (IL) agent consists of two sub-policies: a \textbf{keypose sub-policy} for long-range motion and a \textbf{dense sub-policy} for fine-grained manipulation. Each sub-policy predicts both the next end-effector action and next control mode $m_{t+1} \in \{\texttt{keypose}, \texttt{dense}, \texttt{terminate}\}$, indicating the sub-policy choice for the next action.

\paragraph{Teleoperation.}
We collect data for \acro{} using the iPhone-based interface from~\cite{wu2024tidybot++}, which streams the phone’s real-time 6-DoF pose via the WebXR API and maps it to the robot’s end-effector. Gripper commands are issued through swipe gestures. The WBC from \cref{sec:method-wbc} solves for joint-space actions at each timestep. We record observations and actions at 10\,Hz during teleoperation.

\paragraph{Keypose Sub-policy.}
The keypose sub-policy \(\pi^{\texttt{keypose}}\) handles long-range movements such as reaching, where predicting an absolute end-effector pose provides greater stability than step-wise deltas. It takes as input a third-person point cloud \(\mathcal{P}_t \subset \mathbb{R}^3\), constructed by deprojecting RGB-D images using known intrinsics and extrinsics, and outputs a 6-DoF end-effector pose \(\mathbf{x}^{\textbf{ee}}\), gripper state \(g_{t+1}\), and next control mode \(m_{t+1}\).
Following SPHINX~\cite{sundaresan2024s}, we avoid directly regressing the target end-effector pose. Instead, the policy predicts per-point saliency probabilities over the input cloud and per-point 3D offsets to the ground-truth end-effector position. The point with the highest saliency defines the \emph{salient point}—a task-relevant 3D location such as a keypoint on a cabinet handle (\cref{fig:method}). During training, we supervise offset predictions only at points with high predicted or ground-truth saliency, encouraging the model to focus on task-relevant regions (\cref{fig:method}, shaded offsets). End-effector orientation, gripper state, and control mode are predicted using additional learnable tokens. At test time, we apply the predicted offset at the highest-saliency point to obtain the positional component, and combine it with the predicted orientation and gripper state to form the full end-effector action \(\mathbf{x}^{\text{ee}}\). We then execute interpolated poses from the current pose \(\mathbf{x}^{\text{ee}}_t\) to reach the keypose \(\mathbf{x}^{\text{ee}}\).
Training the keypose policy requires action labels, mode labels, and salient point annotations. For a given dataset of 20 demos, we post-hoc annotate salient points and modes using a lightweight interface, which takes \(\sim\)15 minutes (see \cref{sec:appendix-annotation}). We train the policy using the Transformer-based architecture from SPHINX~\cite{sundaresan2024s}.

\paragraph{Conditioned Keypose Sub-policy.}
\label{sec:homer_cond}
We additionally extend the keypose sub-policy to a salient point-conditioned variant, \acro-\textsc{Cond}, which optionally accepts an externally provided 3D keypoint. This enables us to tap into the internet-scale visual and semantic knowledge encoded in vision-language models (VLMs) by prompting them to localize unseen objects in cluttered scenes, and conditioning \acro-\textsc{Cond} on the resulting deprojected 3D keypoints (\cref{fig:method}). Taking the original point cloud, we first construct a distance-weighted saliency map, where each point’s value is inversely proportional to its distance from the provided keypoint, and the map is normalized to represent probabilities of saliency. We concatenate this saliency map as an additional channel at the input. During training, we apply a masked supervision strategy: in 50\% of samples, the conditioned saliency map is masked out, and the model learns to predict both saliency and actions as in the unconditioned setting; in the remaining 50\%, we pass the unmodified conditioned saliency map and supervise only the action predictions, with offsets penalized only for points with high ground-truth saliency in the conditioned map. This formulation allows the policy to leverage external guidance when available. We further apply visual augmentations during training to promote generalization: adding randomly generated clusters of points to the input point cloud to mimic distractors, and omitting the RGB channel entirely to reduce overfitting to object appearance.

\paragraph{Dense Sub-policy.}  
The dense sub-policy $\pi^{\texttt{dense}}$ is intended for fine-grained manipulation near salient points, such as inserting, aligning, or grasping objects. The input consists of both third-person and wrist-mounted RGB images $\{\mathbf{I}_t^{k} \in \mathbb{R}^{H \times W \times 3}\}_{k=1}^K$, along with the current end-effector state $\mathbf{x}^{\text{ee}}_t \in SE(3)$ computed via forward kinematics from \(\mathbf{q}_t\). The dense policy predicts a 6D delta action $\Delta \mathbf{x}^{\text{ee}}_t \in \mathbb{R}^6$ relative to the current end-effector pose (\cref{fig:method}), as well as the next control mode \(m_{t+1}\). We obtain the target pose as $\mathbf{x}_{t+1}^{\text{ee}} = \mathbf{x}_t^{\text{ee}} + \Delta \mathbf{x}_t^{\text{ee}}$. We instantiate \(\pi^{\texttt{dense}}\) using Diffusion Policy~\cite{chi2023diffusion}, which in practice predicts a horizon of 16 future actions and executes 8 before re-planning, rather than predicting a single delta action at each timestep.

\paragraph{Execution.}
The agent automatically switches between sub-policies based on the \mbox{current} mode \(m_t\):
\vspace{-8px}
\[
(\mathbf{x}^{\textbf{ee}}_{t+1}, m_{t+1}) =
\begin{cases}
\pi^{\texttt{keypose}}(\mathcal{P}_t) & \text{if } m_t = \texttt{keypose} \\
\pi^{\texttt{dense}}(\{\mathbf{I}_t^{k}\}) & \text{if } m_t = \texttt{dense}
\end{cases}
\]
We assume that $m_1$ corresponds to keypose mode, as nearly all manipulation tasks involve first reaching before performing fine-grained manipulation. For each timestep thereafter, the predicted action \(\mathbf{x}^{\textbf{ee}}_{t+1}\) is passed to the WBC to solve for and execute \(\mathbf{q}_{t+1} = \mathcal{W}(\mathbf{x}^{\textbf{ee}}_{t+1})\) (\cref{sec:method-wbc}). \acro~uses the predicted mode \(m_{t+1}\) to select the next sub-policy, enabling dynamic alternation between reaching and manipulation based on learned transitions.

\section{Experiments}
\label{sec:experiments}

\vspace{0.5em}
We deploy \acro{} on the TidyBot++ robot~\cite{wu2024tidybot++}, consisting of a 7-DoF Kinova arm and holonomic base (\cref{fig:hardware_setup}).
With this platform, we evaluate a diverse set of challenging manipulation tasks in both simulation and real-world to investigate three core questions, focusing on the benefits of \acro's imitation learning (IL) agent, whole-body controller (WBC), and generalization capabilities:

\noindent
\textbf{(Q1)}\hspace{0.1cm}\textit{Do hybrid actions help with multi-step tasks combining reaching and fine manipulation?} \linebreak
\textbf{(Q2)}\hspace{0.1cm}\textit{Does the WBC action space improve performance compared to decoupled base-arm actions?} \linebreak
\textbf{(Q3)}\hspace{0.1cm}\textit{Can \acro{} generalize to novel object instances and spatial configurations?}

\subsection{Q1 \& Q2: Are hybrid actions and whole-body control beneficial?}
\label{sec:q1q2-eval}

\paragraph{Baselines.}
We compare \acro{} to baselines varying along two axes: hybrid vs. dense-only action spaces, and whole-body vs. decoupled base-arm control.
Hybrid variants are trained on data annotated post-hoc with control modes and salient points.
WBC baselines are trained from whole-body teleoperation demonstrations (\cref{fig:splash}), while base+arm (B+A) baselines use decoupled teleoperation, in which the base has to be directly teleoperated separately from the arm~\cite{wu2024tidybot++}.

\noindent
\underline{\emph{Diffusion Policy (B+A)}}: A dense policy that predicts 10-DoF relative poses: 3-DoF base pose, 6-DoF end-effector pose, and 1-DoF gripper command. This is comparable to the dense, base-arm policies used in \cite{wu2024tidybot++, prasad2024consistency, fu2024mobile} which notably were trained with 50-200 demos. \\[4pt]
\underline{\acro~\textsc{(B+A)}}: A hybrid policy identical to \acro{}, but predicting either a 3-DoF base keypose, a 6-DoF arm keypose, or a 10-DoF relative pose (as above).\\[4pt]
\underline{\emph{Diffusion Policy (WBC)}}: A dense policy that predicts 7-DoF relative poses: 6-DoF end-effector pose and 1-DoF gripper command, executed through the WBC.\\[4pt]
\underline{\textsc{\acro{} (Ours)}}: A hybrid IL agent that predicts either a 6-DoF end-effector keypose or a 6-DoF relative end-effector pose, plus a 1-DoF gripper command, executed through the WBC.

We expect hybrid action modes and whole-body control (WBC) to each provide advantages in tasks involving wide workspaces, precise phases, and long horizons. \acro~\textsc{(B+A)} and \textsc{DP (WBC)} each capture one of these components, and may perform competitively by partially addressing these challenges. In contrast, \textsc{DP (B+A)}, which lacks both hybrid and whole-body action abstractions, must learn base-arm coordination and long-horizon planning from scratch without structural priors, making the learning problem significantly harder.


\begin{table}[t]
\renewcommand{\arraystretch}{1.1}
\small
\resizebox{\columnwidth}{!}{%
\setlength{\tabcolsep}{3px}
\begin{tabular}{l llcccc}
\toprule
& \textbf{Task} & \textbf{Description} & \textbf{R-R} & \textbf{R-O} & \textbf{P} & \textbf{LH} \\
\midrule
\multirow{3}{*}{\rotatebox[origin=c]{90}{{Sim}}} 
& \emph{Cube} & Pick up cube placed randomly across large workspace & & \textcolor{darkgreen}{\CheckmarkBold} & \textcolor{darkgreen}{\CheckmarkBold} & \\
& \emph{Dishwasher} & Open randomly placed dishwasher door & & \textcolor{darkgreen}{\CheckmarkBold} & \textcolor{darkgreen}{\CheckmarkBold} & \\
& \emph{Cabinet Opening} & Open randomly placed side-hinged cabinet door & & \textcolor{darkgreen}{\CheckmarkBold} & \textcolor{darkgreen}{\CheckmarkBold} & \\
\midrule
\multirow{3}{*}{\rotatebox[origin=c]{90}{{Real}}} 
& \emph{Pillow} & Move pillow placed randomly on carpet to target couch position & \textcolor{darkgreen}{\CheckmarkBold
} & \textcolor{darkgreen}{\CheckmarkBold} & \textcolor{darkgreen}{\CheckmarkBold} & \textcolor{lightgreen}{\CheckmarkBold
} \\
& \emph{TV Remote} & Grasp and open cabinet, retrieve remote, place on stand & \textcolor{darkgreen}{\CheckmarkBold} & \textcolor{lightgreen}{\CheckmarkBold} & \textcolor{darkgreen}{\CheckmarkBold} & \textcolor{darkgreen}{\CheckmarkBold} \\
& \emph{Sweep Trash} & Grasp brush \& sweep at least 3/4 trash clumps into bin. & \textcolor{darkgreen}{\CheckmarkBold} & \textcolor{lightgreen}{\CheckmarkBold} & \textcolor{darkgreen}{\CheckmarkBold} & \textcolor{darkgreen}{\CheckmarkBold} \\
\bottomrule
\end{tabular}
}
\vspace{-5px}
\caption{\small \textbf{Mobile Manipulation Tasks.} We evaluate our approach on 3 simulated and 3 real-world tasks, covering randomized robot poses (R-R), randomized object poses (R-O), need for precision (P), and long-horizon reasoning (LH). Darker checkmarks indicate greater emphasis on the corresponding aspect.}
\label{tab:tasks}
\vspace{-15px}
\end{table}

\begin{figure*}[b]
\raggedleft
\includegraphics[width=0.92\linewidth]{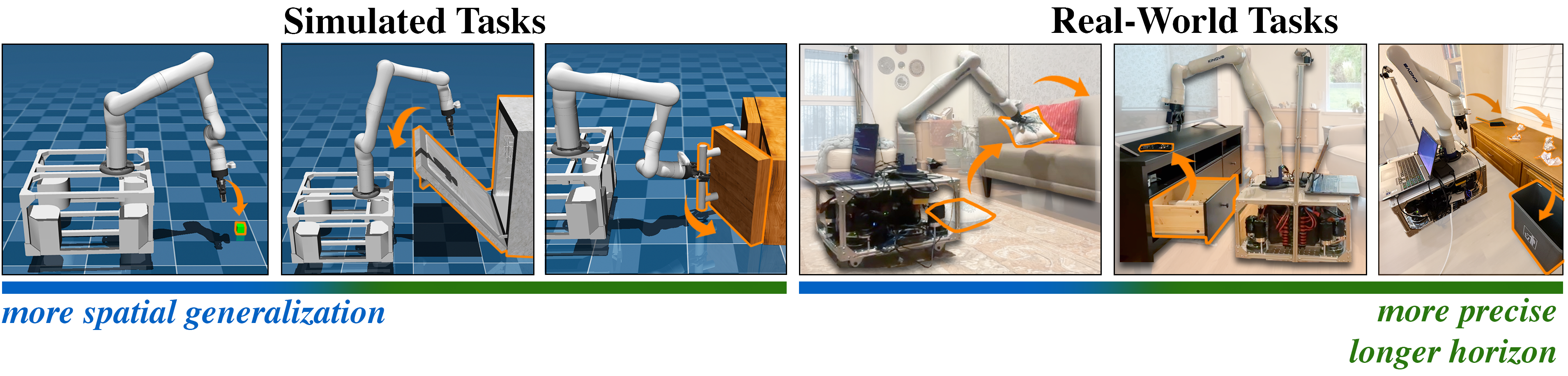}
\\
\vspace{-15px}
\begin{tikzpicture}
\begin{axis}[
    ybar,
    area legend,
    bar width=10pt,
    width=14.5cm,
    height=3.5cm,
    ylabel={Success Rate / 20},
    symbolic x coords={Cube, Dishwasher, Cabinet, Pillow, TV Remote, Sweep Trash},
    xtick=data,
    ymin=0,
    ymax=23,
    clip=false,
    ytick={0,5,10,15,20},
    font=\small, 
    nodes near coords,
    every node near coord/.append style={black},
    axis lines = left,
    enlarge x limits = true,
    legend style={
        at={(0.5,-0.25)},
        anchor=north,
        legend columns=-1,
        draw=none,
        column sep=0pt,
        legend cell align={left},
    }
]
\definecolor{dp_ba}{HTML}{dfe0e2}
\definecolor{dp_wbc}{HTML}{ffd07b}
\definecolor{hybrid_ba}{HTML}{aaaaaa}
\definecolor{homer}{HTML}{fb8b24}

\addplot+[style={fill=dp_ba,draw=none,fill opacity=1.0}] coordinates {(Cube, 5) (Dishwasher, 5) (Cabinet, 9) (Pillow, 5) (TV Remote, 7) (Sweep Trash, 7)};
\addplot+[style={fill=dp_wbc,draw=none,fill opacity=1.0,}] coordinates {(Cube, 9) (Dishwasher, 9) (Cabinet, 11) (Pillow, 8) (TV Remote, 3) (Sweep Trash, 5)};
\addplot+[style={fill=hybrid_ba,draw=none,fill opacity=1.0}] coordinates {(Cube, 10) (Dishwasher, 10) (Cabinet, 12) (Pillow, 10) (TV Remote, 10) (Sweep Trash, 8)};
\addplot+[style={fill=homer,draw=none,fill opacity=1.0}] coordinates {(Cube, 18) (Dishwasher, 13) (Cabinet, 17) (Pillow, 16) (TV Remote, 15) (Sweep Trash, 16)};


\end{axis}
\end{tikzpicture}
\centering                
\begin{tikzpicture}[x=1cm, baseline]   
\definecolor{dp_ba}{HTML}{dfe0e2}
\definecolor{dp_wbc}{HTML}{ffd07b}
\definecolor{hybrid_ba}{HTML}{aaaaaa}
\definecolor{homer}{HTML}{fb8b24}
  \node[draw=none, fill=dp_ba, minimum width=20pt, minimum height=7pt] at (0,0) {};
  \node[anchor=west] at (0.45,0) {\small DP\,(B+A)};

  \node[draw=none, fill=dp_wbc, minimum width=20pt, minimum height=7pt] at (3,0) {};
  \node[anchor=west] at (3.45,0) {\small DP\,(WBC)};

  \node[draw=none, fill=hybrid_ba, minimum width=20pt, minimum height=7pt] at (6,0) {};
  \node[anchor=west] at (6.45,0) {\small HoMeR\,(B+A)};

  \node[draw=none, fill=homer, minimum width=20pt, minimum height=7pt] at (9,0) {};
  \node[anchor=west] at (9.45,0) {\small HoMeR};
\end{tikzpicture}

    \caption{\small \textbf{Benchmarking Results.}
    We evaluate \acro{} on six simulated and real-world tasks (top) that require spatial generalization, precision, and long-horizon reasoning. {\emph{TV Remote}} and {\emph{Sweep Trash}} are particularly challenging due to their multi-step nature. \acro{} consistently outperforms baselines that use only dense actions or decoupled base-arm control, highlighting the benefits of hybrid action modes and whole-body coordination. The performance of all methods is best understood through videos available \href{https://homer-manip.github.io/\#benchmark}{here}.}
    \label{fig:benchmarking_results}
\end{figure*}
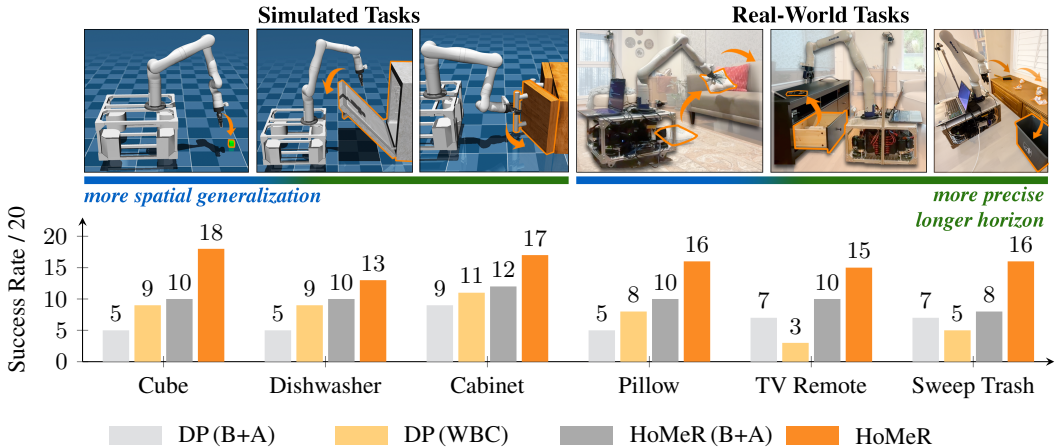

Our benchmark tasks are described in \cref{tab:tasks}.
We train and evaluate all methods using {20 demonstrations} on 3 simulated and 3 real-world tasks.
In \cref{fig:benchmarking_results}, we show illustrations of each task \emph{(top)} and benchmark results \emph{(bottom)}.
Across tasks, DP (B+A) struggles the most with reaching and aligning to targets, particularly in tasks like \emph{Cube}, \emph{Dishwasher}, \emph{Cabinet Opening}, and \emph{Pillow}, with significant randomization in either initial robot poses (R-R) or object placements (R-O).
The dense end-effector deltas output by the policy often veer off course,
leading to failures in reaching pre-manipulation configurations.
DP (WBC) exhibits similar limitations without exploiting keyposes, but performs slightly better. We posit that the simplified end-effector action space enabled by the WBC can be beneficial in the low-data regime. \acro~\textsc{(B+A)} is the strongest baseline, using base and arm keyposes to move to favorable poses before manipulation. This highlights the value of our hybrid IL architecture. However, it struggles when smooth base-arm coordination is required (\emph{Cabinet}, \emph{Dishwasher}, \emph{Sweep Trash}), or when base misalignment affects arm reachability. 

\acro{} achieves the highest success rates across tasks (\cref{fig:benchmarking_results}). In particular, \acro{} is able to perform challenging maneuvers like manipulating appliances larger than the robot itself (\emph{Cabinet}, \emph{Dishwasher}), perform smooth long-horizon motions (\emph{Sweep Trash}), execute precise actions (\emph{TV Remote}), and generalize with randomization in both object poses and initial robot poses. 

\subsection{Q3: Generalization to Novel Scenarios}
\label{sec:q3-generalization}

\setlength{\intextsep}{0pt} \setlength{\columnsep}{8pt}
\begin{wrapfigure}{r}{0.47\textwidth}
\vspace{-20px}
\centering
\includegraphics[width=0.47\textwidth]{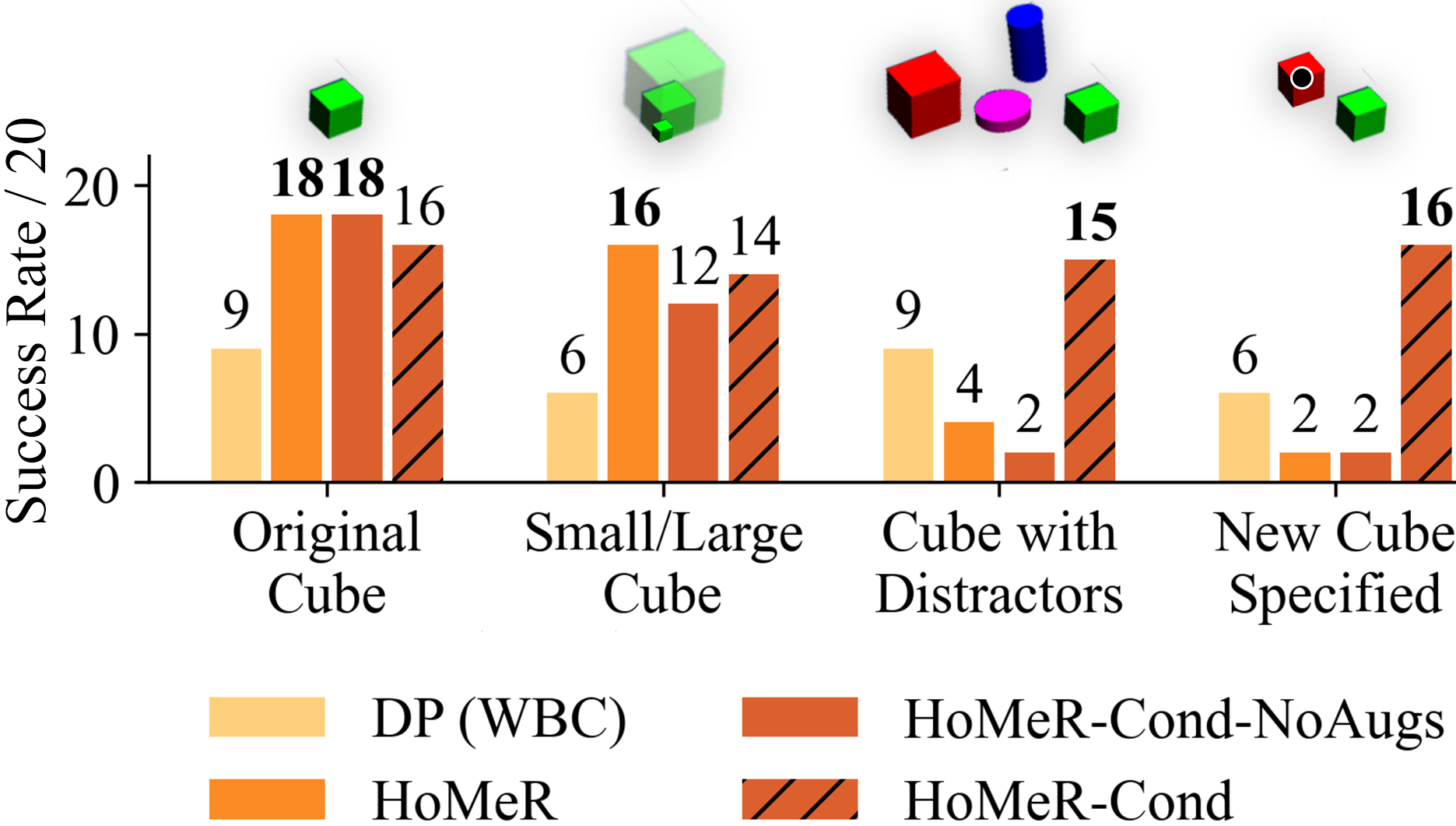}
\caption{\small{\textbf{Generalization Results.} \acro-\textsc{Cond} achieves strong generalization to unseen scenarios by combining salient point conditioning with point cloud augmentations (videos \href{https://homer-manip.github.io/\#generalization}{here}). Without augmentations (\acro-\textsc{Cond}-NoAugs) or conditioning (\acro), performance drops with distractors or novel appearances.}
}
\vspace{2px}
\label{fig:cube_results}
\end{wrapfigure}

To assess generalization, we evaluate \acro-\textsc{Cond}, a variant of \acro{} that (1) conditions the keypose policy on external salient points from a VLM, and (2) trains on point clouds without color and with randomly generated distractor points to improve visual robustness (\cref{sec:homer_cond}). We use MolMo 7B-D~\cite{deitke2024molmo}, a VLM capable of detecting pixel-level keypoints from language prompts (\eg, \cref{fig:method} detect \emph{``cabinet handle''}), and evaluate \acro-\textsc{Cond} on challenging \emph{Cube} variants in simulation: (1) randomizing cube sizes, (2) adding distractors, and (3) retrieving different-colored cubes. \cref{sec:appendix-keypose_cond} details the language prompts used with MolMo and shows qualitative keypoint predictions. Both \acro{} and \acro-\textsc{Cond}-NoAugs perform well in simple settings, but struggle with distractors and novel object appearances. In contrast, \acro-\textsc{Cond} maintains high performance, highlighting the combined importance of salient point conditioning and augmentations for handling clutter and unseen objects.

\subsection{Qualitative Results: Whole-Body Teleoperation in the Wild}

\label{sec:teleop-qualitative}
\begin{figure*}[!htbp]
\vspace{5px}
    \centering
    \makebox[\textwidth][c]{%
        \includegraphics[width=1\linewidth]{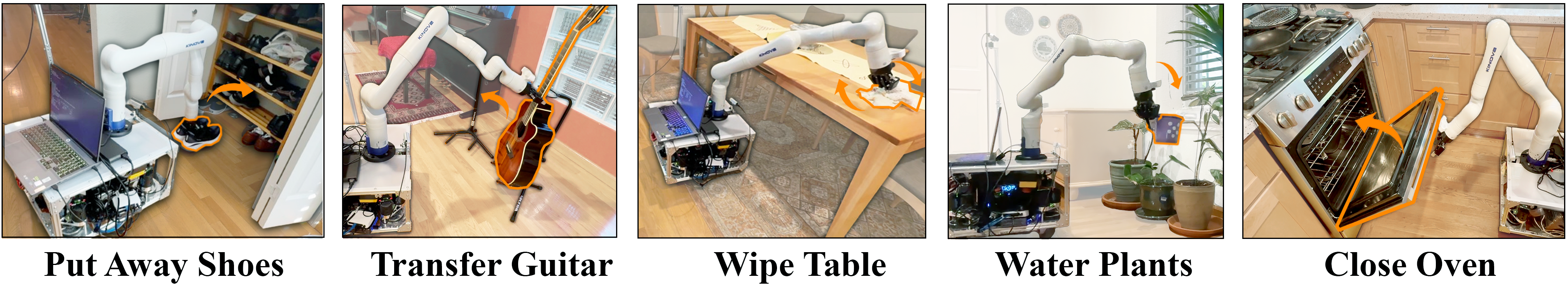}
    }
\vspace{-15px}
\caption{\small \textbf{Teleoperated Tasks:}  We demonstrate a range of teleoperated tasks enabled by our WBC interface, including coordinated whole-body motions and precise behaviors in real household environments.}
\vspace{5px}
\label{fig:teleop_tasks}
\end{figure*}

We also qualitatively assess our WBC through teleoperated demonstrations in a real home. The WBC enables smooth, reliable teleoperation of diverse tasks, including opening and closing cabinets, doors, blinds, and ovens; coordinated motions such as wiping tables and watering plants; and precise maneuvers like putting away shoes or moving a guitar between stands (\cref{fig:teleop_tasks}). The WBC optionally avoids collisions between the arm, base, and camera mounts. These results, best viewed on our \href{https://homer-manip.github.io/#data-collection-video}{website}, highlight our WBC’s potential for scalable in-the-wild teleoperation.

\section{Conclusion}
\label{sec:conclusion}

We present \acro, a hybrid imitation learning framework for mobile manipulators that combines spatially grounded policy learning with a whole-body controller for executing end-effector actions. By switching between keypose and dense control modes, and operating within a lower-dimensional action space, \acro~enables generalizable and precise manipulation. Through real-world evaluations in a real home, we demonstrate that \acro~can perform diverse, everyday tasks with high success rates after training with just 20 demonstrations per task. Our results highlight the benefits of hybrid control and whole-body execution for sample-efficient and generalizable mobile manipulation. We believe \acro~provides a foundation for scalable deployment of assistive robots in real-world, human-centered environments.

\section{Limitations and Future Work}
While this work demonstrates the benefits of whole-body control (WBC) and hybrid action representations for mobile manipulation, several limitations remain, as described below.

\textbf{Collision avoidance.} First, although our WBC accounts for self-collisions and posture regularization, it does not consider collisions with the external environment. Recent works such as~\cite{dalal2024neural} have proposed point-cloud-based collision avoidance for tabletop manipulators, and incorporating similar constraints into our whole-body controller could enable safe operation in tightly constrained or cluttered environments. We leave this integration as an exciting direction for future work. 

\textbf{Active perception.}
We use fixed base-mounted and wrist-mounted camera viewpoints that were manually chosen to cover a wide range of tasks. However, the question of which viewpoints are most useful remains underexplored. With a mobile platform, we are particularly excited about incorporating active perception to select or adapt viewpoints dynamically during task execution.

\textbf{Navigation.}
We note that this work focuses on manipulation and does not address navigation. In practice, navigation is highly complementary and could be interleaved with our manipulation policies to enable truly long-horizon mobile manipulation. 

\textbf{Multi-task.}
Lastly, the policies explored in this work are all single-task and demonstrate promising performance in the limited data regime. In the future, we are excited about scaling \acro~to the multi-task setting, and scaling up teleoperation using our interface. A benefit of our action space is that it simplifies to end-effector actions, which opens up the possibility of co-training on other (both mobile and non-mobile) multi-task datasets in the future.

\footnotesize
\acknowledgments{
Toyota Research Institute, Intrinsic, the Stanford Institute for Human-Centered Artificial Intelligence (HAI) and the Sloan Foundation provided funds to support this work. This work is also in part supported by funds from NSF Awards 2132847, 2327974, and 2006388. Priya Sundaresan is supported by an NSF GRFP. Francis Engelmann is supported by an SNSF PostDoc Mobility Fellowship. 
}
\normalsize

\clearpage

\bibliography{corl}

\begin{thebibliography}{57}
\providecommand{\natexlab}[1]{#1}
\providecommand{\url}[1]{\texttt{#1}}
\expandafter\ifx\csname urlstyle\endcsname\relax
  \providecommand{\doi}[1]{doi: #1}\else
  \providecommand{\doi}{doi: \begingroup \urlstyle{rm}\Url}\fi

\bibitem[Wu et~al.(2023)Wu, Antonova, Kan, Lepert, Zeng, Song, Bohg, Rusinkiewicz, and Funkhouser]{wu2023tidybot}
J.~Wu, R.~Antonova, A.~Kan, M.~Lepert, A.~Zeng, S.~Song, J.~Bohg, S.~Rusinkiewicz, and T.~Funkhouser.
\newblock Tidybot: Personalized robot assistance with large language models.
\newblock \emph{Autonomous Robots}, 2023.

\bibitem[Wu et~al.(2024)Wu, Chong, Holmberg, Prasad, Gao, Khatib, Song, Rusinkiewicz, and Bohg]{wu2024tidybot++}
J.~Wu, W.~Chong, R.~Holmberg, A.~Prasad, Y.~Gao, O.~Khatib, S.~Song, S.~Rusinkiewicz, and J.~Bohg.
\newblock Tidybot++: An open-source holonomic mobile manipulator for robot learning.
\newblock \emph{arXiv preprint arXiv:2412.10447}, 2024.

\bibitem[Brøndmo(2019)]{brondmo2019everyday}
H.~P. Brøndmo.
\newblock Introducing the everyday robot project.
\newblock \url{https://blog.x.company/introducing-the-everyday-robot-project-27860f3461a4}, 2019.
\newblock Accessed: 2025-04-06.

\bibitem[Haviland et~al.(2022)Haviland, S{\"u}nderhauf, and Corke]{haviland2022holistic}
J.~Haviland, N.~S{\"u}nderhauf, and P.~Corke.
\newblock A holistic approach to reactive mobile manipulation.
\newblock \emph{IEEE Robotics and Automation Letters}, 2022.

\bibitem[Nasiriany et~al.(2024)Nasiriany, Maddukuri, Zhang, Parikh, Lo, Joshi, Mandlekar, and Zhu]{nasiriany2024robocasa}
S.~Nasiriany, A.~Maddukuri, L.~Zhang, A.~Parikh, A.~Lo, A.~Joshi, A.~Mandlekar, and Y.~Zhu.
\newblock Robocasa: Large-scale simulation of everyday tasks for generalist robots.
\newblock \emph{arXiv preprint arXiv:2406.02523}, 2024.

\bibitem[Rana et~al.(2023)Rana, Haviland, Garg, Abou-Chakra, Reid, and Suenderhauf]{rana2023sayplan}
K.~Rana, J.~Haviland, S.~Garg, J.~Abou-Chakra, I.~Reid, and N.~Suenderhauf.
\newblock Sayplan: Grounding large language models using 3d scene graphs for scalable robot task planning.
\newblock \emph{arXiv preprint arXiv:2307.06135}, 2023.

\bibitem[Wise et~al.(2016)Wise, Ferguson, King, Diehr, and Dymesich]{wise2016fetch}
M.~Wise, M.~Ferguson, D.~King, E.~Diehr, and D.~Dymesich.
\newblock Fetch and freight: Standard platforms for service robot applications.
\newblock In \emph{Workshop on autonomous mobile service robots}, 2016.

\bibitem[Yamamoto et~al.(2018)Yamamoto, Nishino, Kajima, Ohta, and Ikeda]{yamamoto2018human}
T.~Yamamoto, T.~Nishino, H.~Kajima, M.~Ohta, and K.~Ikeda.
\newblock {Human Support Robot (HSR)}.
\newblock In \emph{ACM SIGGRAPH 2018 emerging technologies}, 2018.

\bibitem[Bjorck et~al.(2025)Bjorck, Castañeda, Cherniadev, Da, Ding, Fan, Fang, Fox, Hu, Huang, Jang, Jiang, Kautz, Kundalia, Lao, Li, Lin, Lin, Liu, Llontop, Magne, Mandlekar, Narayan, Nasiriany, Reed, Tan, Wang, Wang, Wang, Wang, Xiang, Xie, Xu, Xu, Ye, Yu, Zhang, Zhang, Zhao, Zheng, and Zhu]{bjorck2025gr00t}
J.~Bjorck, F.~Castañeda, N.~Cherniadev, X.~Da, R.~Ding, L.~J. Fan, Y.~Fang, D.~Fox, F.~Hu, S.~Huang, J.~Jang, Z.~Jiang, J.~Kautz, K.~Kundalia, L.~Lao, Z.~Li, Z.~Lin, K.~Lin, G.~Liu, E.~Llontop, L.~Magne, A.~Mandlekar, A.~Narayan, S.~Nasiriany, S.~Reed, Y.~L. Tan, G.~Wang, Z.~Wang, J.~Wang, Q.~Wang, J.~Xiang, Y.~Xie, Y.~Xu, Z.~Xu, S.~Ye, Z.~Yu, A.~Zhang, H.~Zhang, Y.~Zhao, R.~Zheng, and Y.~Zhu.
\newblock Gr00t n1: An open foundation model for generalist humanoid robots.
\newblock \emph{arXiv preprint arXiv:2503.14734}, 2025.

\bibitem[Cheng et~al.(2024)Cheng, Ji, Chen, Yang, Yang, and Wang]{cheng2024expressive}
X.~Cheng, Y.~Ji, J.~Chen, R.~Yang, G.~Yang, and X.~Wang.
\newblock {Expressive Whole-body Control for Humanoid Robots}.
\newblock \emph{arXiv preprint arXiv:2402.16796}, 2024.

\bibitem[Qiu et~al.(2025)Qiu, Yang, Cheng, Chawla, Li, He, Yan, Paulsen, Yang, Yi, et~al.]{qiu2025humanoid}
R.-Z. Qiu, S.~Yang, X.~Cheng, C.~Chawla, J.~Li, T.~He, G.~Yan, L.~Paulsen, G.~Yang, S.~Yi, et~al.
\newblock Humanoid policy\~{} human policy.
\newblock \emph{arXiv preprint arXiv:2503.13441}, 2025.

\bibitem[Fu et~al.(2024)Fu, Zhao, Wu, Wetzstein, and Finn]{fu2024humanplus}
Z.~Fu, Q.~Zhao, Q.~Wu, G.~Wetzstein, and C.~Finn.
\newblock Humanplus: Humanoid shadowing and imitation from humans.
\newblock \emph{arXiv preprint arXiv:2406.10454}, 2024.

\bibitem[{1X Technologies}(2025)]{1x2025worldmodel}
{1X Technologies}.
\newblock 1x world model challenge for humanoid robots, 2025.
\newblock URL \url{https://github.com/1x-technologies/1xgpt}.

\bibitem[Lu et~al.(2025)Lu, Cheng, Li, Yang, Ji, Yuan, Yang, Yi, and Wang]{wang2025mobile}
C.~Lu, X.~Cheng, J.~Li, S.~Yang, M.~Ji, C.~Yuan, G.~Yang, S.~Yi, and X.~Wang.
\newblock {Mobile-TeleVision: Predictive Motion Priors for Humanoid Whole-Body Control}.
\newblock In \emph{{Proc. {IEEE} Int. Conf. Robotics and Automation (ICRA)}}, 2025.

\bibitem[Ji et~al.(2024)Ji, Peng, Liu, Cheng, Yang, Yang, and Wang]{wang2024exbody2}
M.~Ji, X.~Peng, F.~Liu, X.~Cheng, R.~Yang, G.~Yang, and X.~Wang.
\newblock Exbody2: Advanced expressive humanoid whole-body control.
\newblock \emph{arXiv preprint arXiv:2412.13196}, 2024.

\bibitem[{Unitree Robotics}(2025)]{unitree2025h1}
{Unitree Robotics}.
\newblock Unitree h1: Full-size universal humanoid robot, 2025.
\newblock URL \url{https://www.unitree.com/h1}.

\bibitem[Sferrazza et~al.(2024)Sferrazza, Huang, Lin, Lee, and Abbeel]{sferrazza2024humanoidbench}
C.~Sferrazza, D.-M. Huang, X.~Lin, Y.~Lee, and P.~Abbeel.
\newblock {HumanoidNench: Simulated Humanoid Benchmark for Whole-body Locomotion and Manipulation}.
\newblock \emph{arXiv preprint arXiv:2403.10506}, 2024.

\bibitem[Shah et~al.(2022)Shah, Osinski, Ichter, and Levine]{shah2022lmnav}
D.~Shah, B.~Osinski, B.~Ichter, and S.~Levine.
\newblock {LM-Nav: Robotic Navigation with Large Pre-Trained Models of Language, Vision, and Action}.
\newblock \emph{arXiv preprint arXiv:2207.04429}, 2022.

\bibitem[Lee et~al.(2020)Lee, Hwangbo, Wellhausen, Koltun, and Hutter]{lee2020learning}
J.~Lee, J.~Hwangbo, L.~Wellhausen, V.~Koltun, and M.~Hutter.
\newblock {Learning Quadrupedal Locomotion over Challenging Terrain}.
\newblock \emph{Science Robotics}, 2020.

\bibitem[Kumar et~al.(2021)Kumar, Xu, Pathak, and Malik]{kumar2021rma}
A.~Kumar, Z.~Xu, D.~Pathak, and J.~Malik.
\newblock {RMA: Rapid Motor Adaptation for Legged Robots}.
\newblock \emph{arXiv preprint arXiv:2107.04034}, 2021.

\bibitem[Agarwal et~al.(2023)Agarwal, Kumar, Malik, and Pathak]{agarwal2023legged}
A.~Agarwal, A.~Kumar, J.~Malik, and D.~Pathak.
\newblock Legged locomotion in challenging terrains using egocentric vision.
\newblock In \emph{Conference on robot learning}, 2023.

\bibitem[Margolis et~al.(2022)Margolis, Chen, Paigwar, Fu, Kim, Kim, and Agrawal]{margolis2022learning}
G.~B. Margolis, T.~Chen, K.~Paigwar, X.~Fu, D.~Kim, S.~Kim, and P.~Agrawal.
\newblock {Learning to Jump from Pixels}.
\newblock \emph{Proceedings of the 5th Conference on Robot Learning (CoRL)}, 2022.

\bibitem[Ha et~al.(2024)Ha, Gao, Fu, Tan, and Song]{ha2024umi}
H.~Ha, Y.~Gao, Z.~Fu, J.~Tan, and S.~Song.
\newblock Umi on legs: Making manipulation policies mobile with manipulation-centric whole-body controllers.
\newblock In \emph{arXiv preprint arXiv:2407.10353}, 2024.

\bibitem[Fu et~al.(2022)Fu, Cheng, and Pathak]{fu2022deep}
Z.~Fu, X.~Cheng, and D.~Pathak.
\newblock Deep whole-body control: Learning a unified policy for manipulation and locomotion.
\newblock In \emph{Conference on Robot Learning ({CoRL})}, 2022.

\bibitem[Liu et~al.(2024)Liu, Chen, Cheng, Ji, Qiu, Yang, and Wang]{liu2024visual}
M.~Liu, Z.~Chen, X.~Cheng, Y.~Ji, R.~Qiu, R.~Yang, and X.~Wang.
\newblock Visual whole-body control for legged loco-manipulation.
\newblock \emph{The 8th Conference on Robot Learning}, 2024.

\bibitem[Br\"udigam et~al.(2024)Br\"udigam, Abbas, Sorokin, Fang, Hung, Guru, Sosnowski, Wang, Hirche, and Le~Cleac'h]{bruedigam2024jacta}
J.~Br\"udigam, A.~A. Abbas, M.~Sorokin, K.~Fang, B.~Hung, M.~Guru, S.~Sosnowski, J.~Wang, S.~Hirche, and S.~Le~Cleac'h.
\newblock Jacta: A versatile planner for learning dexterous and whole-body manipulation.
\newblock \emph{arXiv preprint arXiv:2408.01258}, 2024.

\bibitem[Sundaresan et~al.(2024)Sundaresan, Hu, Vuong, Bohg, and Sadigh]{sundaresan2024s}
P.~Sundaresan, H.~Hu, Q.~Vuong, J.~Bohg, and D.~Sadigh.
\newblock {What's the Move? Hybrid Imitation Learning via Salient Points}.
\newblock \emph{Proc. Int. Conf. on Learning Representations}, 2024.

\bibitem[Shi et~al.(2023)Shi, Sharma, Zhao, and Finn]{shi2023waypoint}
L.~X. Shi, A.~Sharma, T.~Z. Zhao, and C.~Finn.
\newblock {Waypoint-Based Imitation Learning for Robotic Manipulation}.
\newblock \emph{Conference on Robot Learning (CoRL)}, 2023.

\bibitem[Belkhale et~al.(2023)Belkhale, Cui, and Sadigh]{belkhale2023hydra}
S.~Belkhale, Y.~Cui, and D.~Sadigh.
\newblock {Hydra: Hybrid Robot Actions for Imitation Learning}.
\newblock In \emph{Conference on Robot Learning (CoRL)}, 2023.

\bibitem[Shridhar et~al.(2023)Shridhar, Manuelli, and Fox]{shridhar2023perceiver}
M.~Shridhar, L.~Manuelli, and D.~Fox.
\newblock {Perceiver-Actor: A Multi-task Transformer for Robotic Manipulation}.
\newblock In \emph{Conference on Robot Learning (CoRL)}, 2023.

\bibitem[James and Davison(2022)]{james2022q}
S.~James and A.~J. Davison.
\newblock Q-attention: Enabling efficient learning for vision-based robotic manipulation.
\newblock \emph{IEEE Robotics and Automation Letters}, 7\penalty0 (2):\penalty0 1612--1619, 2022.

\bibitem[Hutter et~al.(2017)Hutter, Gehring, Lauber, Gunther, Bellicoso, Tsounis, Fankhauser, Diethelm, Bachmann, Bl{\"o}sch, et~al.]{hutter2017anymal}
M.~Hutter, C.~Gehring, A.~Lauber, F.~Gunther, C.~D. Bellicoso, V.~Tsounis, P.~Fankhauser, R.~Diethelm, S.~Bachmann, M.~Bl{\"o}sch, et~al.
\newblock {ANYMal - Toward Legged Robots for Harsh Environments}.
\newblock \emph{Advanced Robotics}, 2017.

\bibitem[Ferrolho et~al.(2023)Ferrolho, Ivan, Merkt, Havoutis, and Vijayakumar]{ferrolho2023roloma}
H.~Ferrolho, V.~Ivan, W.~Merkt, I.~Havoutis, and S.~Vijayakumar.
\newblock Roloma: Robust loco-manipulation for quadruped robots with arms.
\newblock \emph{Autonomous Robots}, 47\penalty0 (8):\penalty0 1463--1481, 2023.

\bibitem[Fu et~al.(2024)Fu, Zhao, and Finn]{fu2024mobile}
Z.~Fu, T.~Z. Zhao, and C.~Finn.
\newblock {Mobile ALOHA: Learning Bimanual Mobile Manipulation with Low-Cost Whole-Body Teleoperation}.
\newblock In \emph{Conference on Robot Learning (CoRL)}, 2024.

\bibitem[Brohan et~al.(2022)Brohan, Brown, Carbajal, Chebotar, Dabis, Finn, Gopalakrishnan, Hausman, Herzog, Hsu, et~al.]{brohan2022rt}
A.~Brohan, N.~Brown, J.~Carbajal, Y.~Chebotar, J.~Dabis, C.~Finn, K.~Gopalakrishnan, K.~Hausman, A.~Herzog, J.~Hsu, et~al.
\newblock Rt-1: Robotics transformer for real-world control at scale.
\newblock \emph{arXiv preprint arXiv:2212.06817}, 2022.

\bibitem[Hirose et~al.(2024)Hirose, Glossop, Sridhar, Shah, Mees, and Levine]{hirose2024lelan}
N.~Hirose, C.~Glossop, A.~Sridhar, D.~Shah, O.~Mees, and S.~Levine.
\newblock {LeLan: Learning a Language-conditioned Navigation Policy from In-the-wild Videos}.
\newblock \emph{Conference on Robot Learning (CoRL)}, 2024.

\bibitem[Yokoyama et~al.(2024)Yokoyama, Ha, Batra, Wang, and Bucher]{yokoyama2024vlfm}
N.~Yokoyama, S.~Ha, D.~Batra, J.~Wang, and B.~Bucher.
\newblock {VLFM: Vision-language Frontier Maps for Zero-shot Semantic Navigation}.
\newblock In \emph{{Proc. {IEEE} Int. Conf. Robotics and Automation (ICRA)}}, 2024.

\bibitem[Jauhri et~al.(2024)Jauhri, Lueth, and Chalvatzaki]{jauhri2024active}
S.~Jauhri, S.~Lueth, and G.~Chalvatzaki.
\newblock {Active-perceptive Motion Generation for Mobile Manipulation}.
\newblock In \emph{{Proc. {IEEE} Int. Conf. Robotics and Automation (ICRA)}}, 2024.

\bibitem[Schaal(1996)]{schaal1996learning}
S.~Schaal.
\newblock Learning from demonstration.
\newblock \emph{Advances in neural information processing systems}, 1996.

\bibitem[Ross et~al.(2011)Ross, Gordon, and Bagnell]{ross2011reduction}
S.~Ross, G.~Gordon, and D.~Bagnell.
\newblock A reduction of imitation learning and structured prediction to no-regret online learning.
\newblock In \emph{Proceedings of the 14th International Conference on Artificial Intelligence and Statistics (AISTATS)}, 2011.

\bibitem[Levine et~al.(2016)Levine, Finn, Darrell, and Abbeel]{levine2016end}
S.~Levine, C.~Finn, T.~Darrell, and P.~Abbeel.
\newblock End-to-end training of deep visuomotor policies.
\newblock In \emph{Proceedings of the 29th IEEE Conference on Computer Vision and Pattern Recognition (CVPR)}, 2016.

\bibitem[Chi et~al.(2023)Chi, Xu, Feng, Cousineau, Du, Burchfiel, Tedrake, and Song]{chi2023diffusion}
C.~Chi, Z.~Xu, S.~Feng, E.~Cousineau, Y.~Du, B.~Burchfiel, R.~Tedrake, and S.~Song.
\newblock {Diffusion Policy: Visuomotor Policy Learning via Action Diffusion}.
\newblock \emph{The International Journal of Robotics Research}, 2023.

\bibitem[Zhao et~al.(2023)Zhao, Kumar, Levine, and Finn]{zhao2023learning}
T.~Z. Zhao, V.~Kumar, S.~Levine, and C.~Finn.
\newblock {Learning Fine-grained Bimanual Manipulation with Low-cost Hardware}.
\newblock In \emph{Proc. Robotics: Science and Systems (RSS)}, 2023.

\bibitem[Team et~al.(2025)Team, Abeyruwan, Ainslie, Alayrac, Arenas, Armstrong, Balakrishna, Baruch, Bauza, Blokzijl, et~al.]{team2025gemini}
G.~R. Team, S.~Abeyruwan, J.~Ainslie, J.-B. Alayrac, M.~G. Arenas, T.~Armstrong, A.~Balakrishna, R.~Baruch, M.~Bauza, M.~Blokzijl, et~al.
\newblock {Gemini Robotics: Bringing AI into the Physical World}.
\newblock \emph{arXiv preprint arXiv:2503.20020}, 2025.

\bibitem[Black et~al.(2024)Black, Brown, Driess, Esmail, Equi, Finn, Fusai, Groom, Hausman, Ichter, et~al.]{black2024pi_0}
K.~Black, N.~Brown, D.~Driess, A.~Esmail, M.~Equi, C.~Finn, N.~Fusai, L.~Groom, K.~Hausman, B.~Ichter, et~al.
\newblock {$pi\_0$: A Vision-Language-Action Flow Model for General Robot Control}.
\newblock \emph{arXiv preprint arXiv:2410.24164}, 2024.

\bibitem[Collaboration et~al.(2023)Collaboration, O'Neill, Rehman, Gupta, Maddukuri, Gupta, et~al.]{open_x_embodiment_rt_x_2023}
O.~X.-E. Collaboration, A.~O'Neill, A.~Rehman, A.~Gupta, A.~Maddukuri, A.~Gupta, et~al.
\newblock {Open {X-E}mbodiment: Robotic Learning Datasets and {RT-X} Models}.
\newblock \url{https://arxiv.org/abs/2310.08864}, 2023.

\bibitem[Kim et~al.(2024)Kim, Pertsch, Karamcheti, Xiao, Balakrishna, Nair, Rafailov, Foster, Lam, Sanketi, Vuong, Kollar, Burchfiel, Tedrake, Sadigh, Levine, Liang, and Finn]{kim24openvla}
M.~Kim, K.~Pertsch, S.~Karamcheti, T.~Xiao, A.~Balakrishna, S.~Nair, R.~Rafailov, E.~Foster, G.~Lam, P.~Sanketi, Q.~Vuong, T.~Kollar, B.~Burchfiel, R.~Tedrake, D.~Sadigh, S.~Levine, P.~Liang, and C.~Finn.
\newblock {OpenVLA: An Open-Source Vision-Language-Action Model}.
\newblock \emph{arXiv preprint arXiv:2406.09246}, 2024.

\bibitem[Team et~al.(2024)Team, Ghosh, Walke, Pertsch, Black, Mees, Dasari, Hejna, Kreiman, Xu, et~al.]{team2024octo}
O.~M. Team, D.~Ghosh, H.~Walke, K.~Pertsch, K.~Black, O.~Mees, S.~Dasari, J.~Hejna, T.~Kreiman, C.~Xu, et~al.
\newblock Octo: An open-source generalist robot policy.
\newblock \emph{arXiv preprint arXiv:2405.12213}, 2024.

\bibitem[Goyal et~al.(2023)Goyal, Xu, Guo, Blukis, Chao, and Fox]{goyal2023rvt}
A.~Goyal, J.~Xu, Y.~Guo, V.~Blukis, Y.-W. Chao, and D.~Fox.
\newblock {RVT: Robotic View Transformer for 3D Object Manipulation}.
\newblock In \emph{Conference on Robot Learning (CoRL)}, 2023.

\bibitem[Goyal et~al.(2024)Goyal, Blukis, Xu, Guo, Chao, and Fox]{goyal2024rvt}
A.~Goyal, V.~Blukis, J.~Xu, Y.~Guo, Y.-W. Chao, and D.~Fox.
\newblock {RVT-2: Learning Precise Manipulation from Few Demonstrations}.
\newblock \emph{Proc. Robotics: Science and Systems (RSS)}, 2024.

\bibitem[Sundaresan et~al.(2023)Sundaresan, Belkhale, Sadigh, and Bohg]{sundaresan2023kite}
P.~Sundaresan, S.~Belkhale, D.~Sadigh, and J.~Bohg.
\newblock {KITE: Keypoint-Conditioned Policies for Semantic Manipulation}.
\newblock In \emph{Conference on Robot Learning}, 2023.

\bibitem[Todorov(2012)]{todorov2012mujoco}
E.~Todorov.
\newblock {MuJoCo: A Physics Engine for Model-Based Control}.
\newblock \url{http://www.mujoco.org}, 2012.
\newblock Accessed: 2025-04-15.

\bibitem[Zakka(2024)]{mink}
K.~Zakka.
\newblock {Mink: Python inverse kinematics based on MuJoCo}, July 2024.
\newblock URL \url{https://github.com/kevinzakka/mink}.

\bibitem[Murray et~al.(1994)Murray, Li, and Sastry]{murray1994mathematical}
R.~M. Murray, Z.~Li, and S.~S. Sastry.
\newblock \emph{A Mathematical Introduction to Robotic Manipulation}.
\newblock CRC Press, 1994.

\bibitem[Prasad et~al.(2024)Prasad, Lin, Wu, Zhou, and Bohg]{prasad2024consistency}
A.~Prasad, K.~Lin, J.~Wu, L.~Zhou, and J.~Bohg.
\newblock Consistency policy: Accelerated visuomotor policies via consistency distillation.
\newblock \emph{arXiv preprint arXiv:2405.07503}, 2024.

\bibitem[Deitke et~al.(2024)Deitke, Clark, Lee, Tripathi, Yang, Park, Salehi, Muennighoff, Lo, Soldaini, et~al.]{deitke2024molmo}
M.~Deitke, C.~Clark, S.~Lee, R.~Tripathi, Y.~Yang, J.~S. Park, M.~Salehi, N.~Muennighoff, K.~Lo, L.~Soldaini, et~al.
\newblock Molmo and pixmo: Open weights and open data for state-of-the-art multimodal models.
\newblock \emph{arXiv preprint arXiv:2409.17146}, 2024.

\bibitem[Dalal et~al.(2024)Dalal, Yang, Mendonca, Khaky, Salakhutdinov, and Pathak]{dalal2024neural}
M.~Dalal, J.~Yang, R.~Mendonca, Y.~Khaky, R.~Salakhutdinov, and D.~Pathak.
\newblock Neural mp: A generalist neural motion planner.
\newblock \emph{arXiv preprint arXiv:2409.05864}, 2024.

\end{thebibliography}
\clearpage
\newpage
\appendix
\begin{Large}
\begin{center}
\textbf{Appendix: Learning In-the-Wild Mobile Manipulation via \\ Hybrid Imitation and Whole-Body Control}
\end{center}
\end{Large}

\section{Whole-Body Controller Implementation Details}
\label{sec:appendix-wbc}

We implement the whole-body controller (WBC) described in \cref{sec:method-wbc} using MuJoCo~\cite{todorov2012mujoco} and the mink inverse kinematics library~\cite{mink}. 

\paragraph{Model and Tasks.}
We load the MuJoCo model of the robot, with two camera mounts attached to the base, from an MJCF file. The WBC includes the following tasks:
\begin{itemize}
    \item \textbf{End-effector pose task:} A 6-DoF frame task is defined at the end-effector, with cost weights $W_{\text{ee}} = 1.0$ for both position and orientation tracking.
    \item \textbf{Posture task:} A quadratic penalty encourages the robot to remain near a neutral joint configuration. We set $W_{\text{posture}} = 2 \times 10^{-3}$ for all non-base joints. The target configuration corresponds to a tucked arm posture used across all tasks.
    \item \textbf{Base damping task:} A damping cost of $W_{\text{damping}} = 1.5$ is applied to base velocities to prevent excessive motion.
\end{itemize}

\paragraph{Constraints.}
We enforce the following limits during IK:
\begin{itemize}
    \item \textbf{Velocity limits:} Base velocities are capped at $(0.5, 0.5, \pi/2)$ m/s and rad/s. Arm joints are limited to approximately $80^\circ$/s for the first four joints and $140^\circ$/s for the wrist joints.
    \item \textbf{Joint position limits:} All joint limits of the robot are enforced.
    \item \textbf{(Optional) Collision limits:} We define geometric collision pairs between the arm, gripper, base, and camera mounts, with a 2\,cm safety margin and 10\,cm detection range. This constraint was not needed in our benchmarking experiments (\cref{fig:benchmarking_results}) as our placement of cameras was not at high collision risk with the whole-body motions, but remains available as a flexible add-on and is demonstrated during teleoperation (\cref{fig:teleop_tasks}).

\end{itemize}

\paragraph{Solver parameters.}
We solve the IK problem using mink's QP solver with a Levenberg-Marquardt damping factor of 1.0. The solver runs for up to 20 iterations with a convergence threshold of $10^{-4}$ for both position and orientation errors. Joint velocities are integrated using Euler integration.

\paragraph{Usage.}
At runtime, the solver takes as input a desired end-effector pose and the current joint configuration, and returns a joint position command by solving the constrained IK problem and integrating the resulting joint velocities. All weights and thresholds are fixed and reused across all tasks without any per-task tuning.

\section{Hybrid IL Implementation Details}

\subsection{Keypose Policy}
\label{sec:appendix-keypose}

We implement the keypose policy using a Transformer that operates on point clouds to predict a 6-DoF end-effector pose. The policy first classifies per-point saliency and then regresses a per-point offset to the target end-effector position. Rotation (as quaternions), gripper state, and control mode are predicted using additional learnable tokens. The network architecture uses 6 Transformer layers with 512-dimensional embeddings and 8 attention heads. No positional encodings are used, as the point cloud input is unordered.

Following \cite{sundaresan2024s}, the full training objective is a simple unweighted sum of the following: (1) salient point classification loss, (2) offset regression loss on high-saliency points, (3) MSE on normalized quaternions, (4) binary cross-entropy on gripper state, and (5) cross-entropy loss on control mode.

 We apply temporal augmentation by including intermediate steps from the controller’s motion trajectory toward each annotated keypose. For each waypoint segment, we train not only on the initial observation but also on a prefix of the interpolated segment. We use $\alpha = 0.2$, meaning we sample the first 20\% of timesteps in the segment. This increases the data sixfold in most cases and improves performance across tasks.

Additionally, we apply spatial augmentations by randomly translating the entire point cloud and corresponding action label within a 5\,cm cube. No vision-based pre-processing or segmentation is used beyond cropping to workspace bounds. We train for 2000 epochs using Adam with a base learning rate of $1e^{-4}$ and cosine decay, gradient clipping (max norm 1), dropout of 0.1, batch size 64, and exponential moving average (EMA) with decay annealed up to 0.9999. All evaluations use the final checkpoint.

\subsubsection{Data Annotation}
\label{sec:appendix-annotation}
\begin{figure}[!htbp]
    \centering
    {\includegraphics[width=1.0\linewidth]{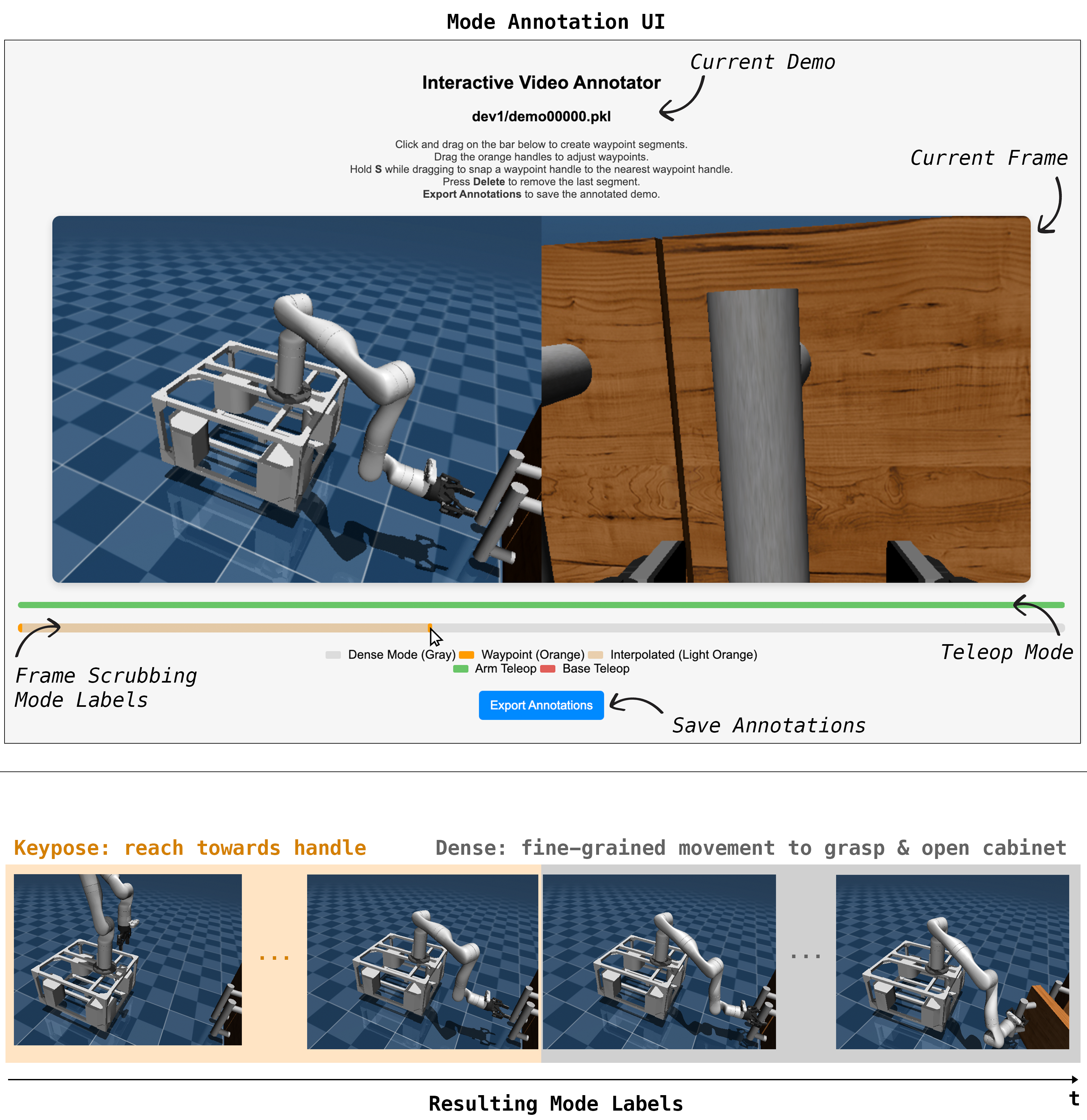}}
    \caption{\textbf{\acro~Mode Annotation Example:}
To train \acro, we annotate control modes using a custom UI that supports frame-by-frame scrubbing and Shift + Click/Drag segmentation. For the \emph{Cabinet} task shown above, we label the reaching motion as keypose (orange) and the grasping and opening phase as dense (gray). A demonstration of this annotation process is available \href{https://homer-manip.github.io/\#mode\_annot}{here}.}

    \label{fig:mode_interface}
\end{figure}

\begin{figure}[!htbp]
    \centering
    {\includegraphics[width=1.0\linewidth]{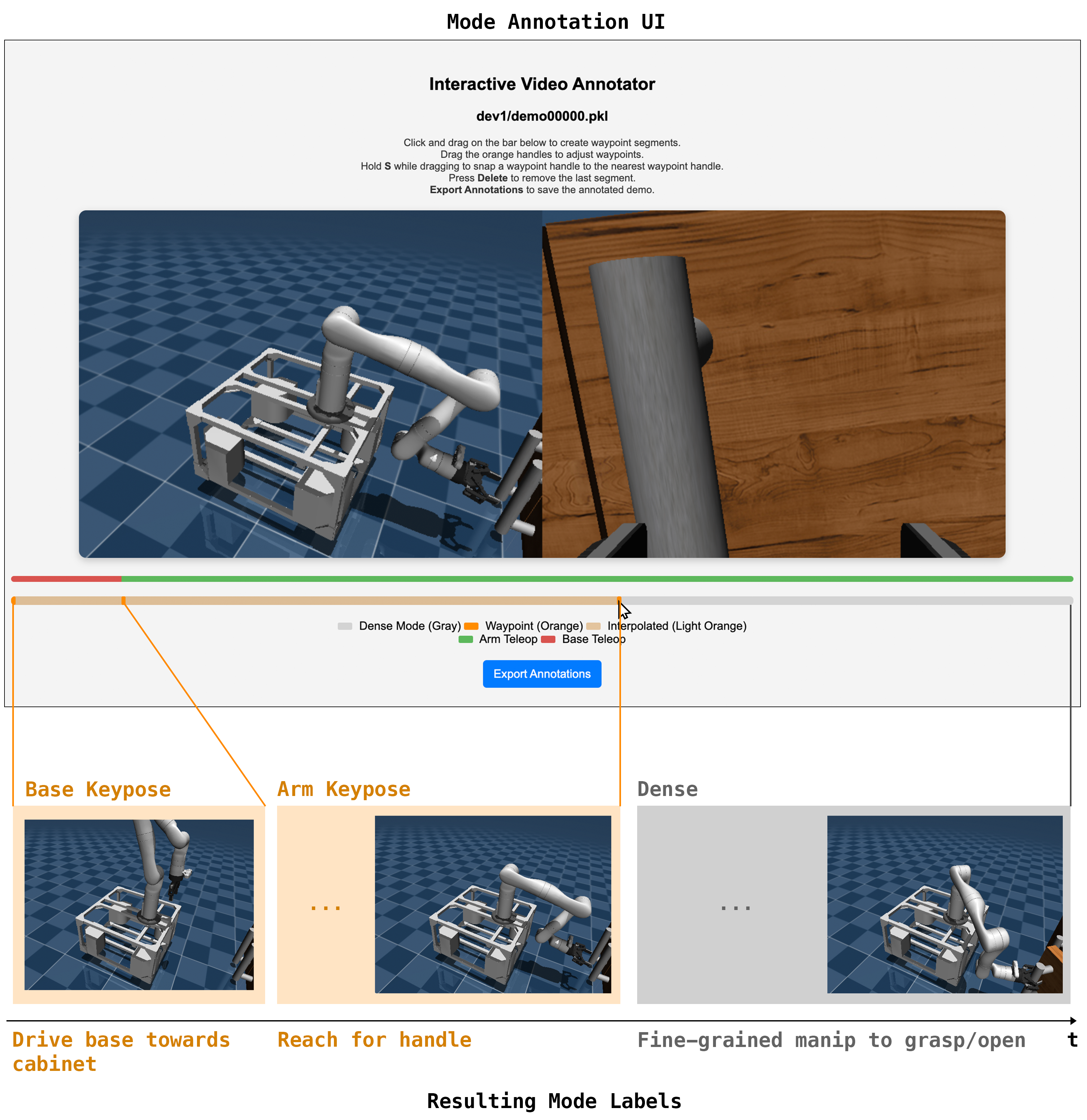}}
    \caption{\textbf{\acro~(B+A) Mode Annotation Example:}
Above we visualize mode annotation on a  \emph{Cabinet} demo collected with separate base+arm teleoperation, with which to train \acro ~(B+A). The demonstration consists of first driving the base towards the cabinet (keypose), reaching the arm towards the handle (keypose), and finally grasping and opening the cabinet (dense). We visualize the base (red) or arm (green) control mode for more intuitive labeling.}

    \label{fig:mode_interface_homerba}
\end{figure}

\begin{figure}[!htbp]
    \centering
    \setlength{\fboxsep}{0pt}
    \fcolorbox{black}{white}{\includegraphics[width=0.5\linewidth]{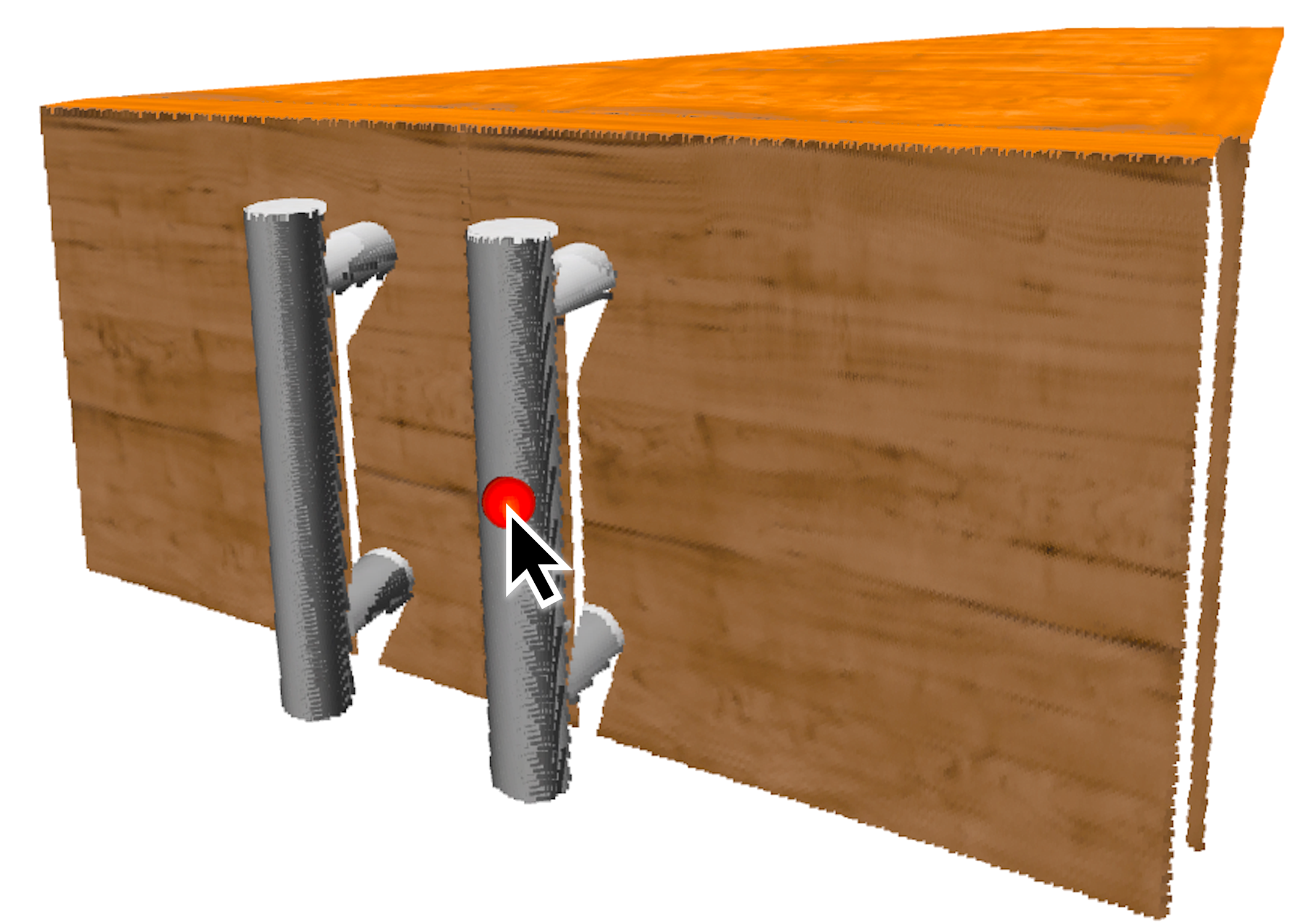}}
    \caption{\textbf{Salient Point Annotation Interface.}
    For frames labeled as keypose using the Mode Annotation Interface \cref{fig:mode_interface}, demonstrators can specify a task-relevant salient point (i.e. the handle in the \emph{Cabinet} task) by simply clicking a desired point within the reconstructed point cloud (video \href{https://homer-manip.github.io/\#sp-labeling-video}{here}). Together, these lightweight interfaces provide all the additional supervision necessary to train the keypose policy.}
    \label{fig:salient_point_interface}
\end{figure}

Training the keypose policy requires labels for modes and salient points. We provide these annotations on teleoperated demonstrations using a lightweight custom interface.  As shown in \cref{fig:mode_interface} and \cref{fig:mode_interface_homerba}, annotators first segment each demonstration into keypose and dense control modes by clicking and dragging on a timeline. For frames labeled as keypose, annotators then specify a salient point by clicking on a task-relevant location in the 3D point cloud interface (\cref{fig:salient_point_interface}). Each demonstration typically contains 1--3 such annotations, and full annotation of a 20-demo dataset takes around 15 minutes. These labels supervise both the saliency classification and offset regression components of the keypose policy.

\vspace{0.2cm}
\vspace{0.2cm}
\vspace{0.1cm}

\newpage
\subsubsection{Salient-Point Conditioned Keypose Policy}
\label{sec:appendix-keypose_cond}

To improve robustness and generalization, we extend the keypose policy to accept externally specified salient points rather than learning to predict them from scratch. These points are encoded as a soft saliency map over the input point cloud and allow the keypose model to attend to a pre-specified point.

We train this variant with a masked supervision strategy. 50\% of the time, we include the saliency map, and the policy learns to predict actions relative to given salient points when available. In the other 50\%, we mask out the saliency map, and the model learns to predict the map in addition to the action, in order to encourage the model to learn useful features of the input point cloud. We also apply data augmentations to the input point clouds, including removal of color channels and injection of distractor points, to improve visual robustness in cluttered scenes.

\vspace{0.3cm}
\begin{figure*}[!htbp]
    \centering {\includegraphics[width=1.0\linewidth]{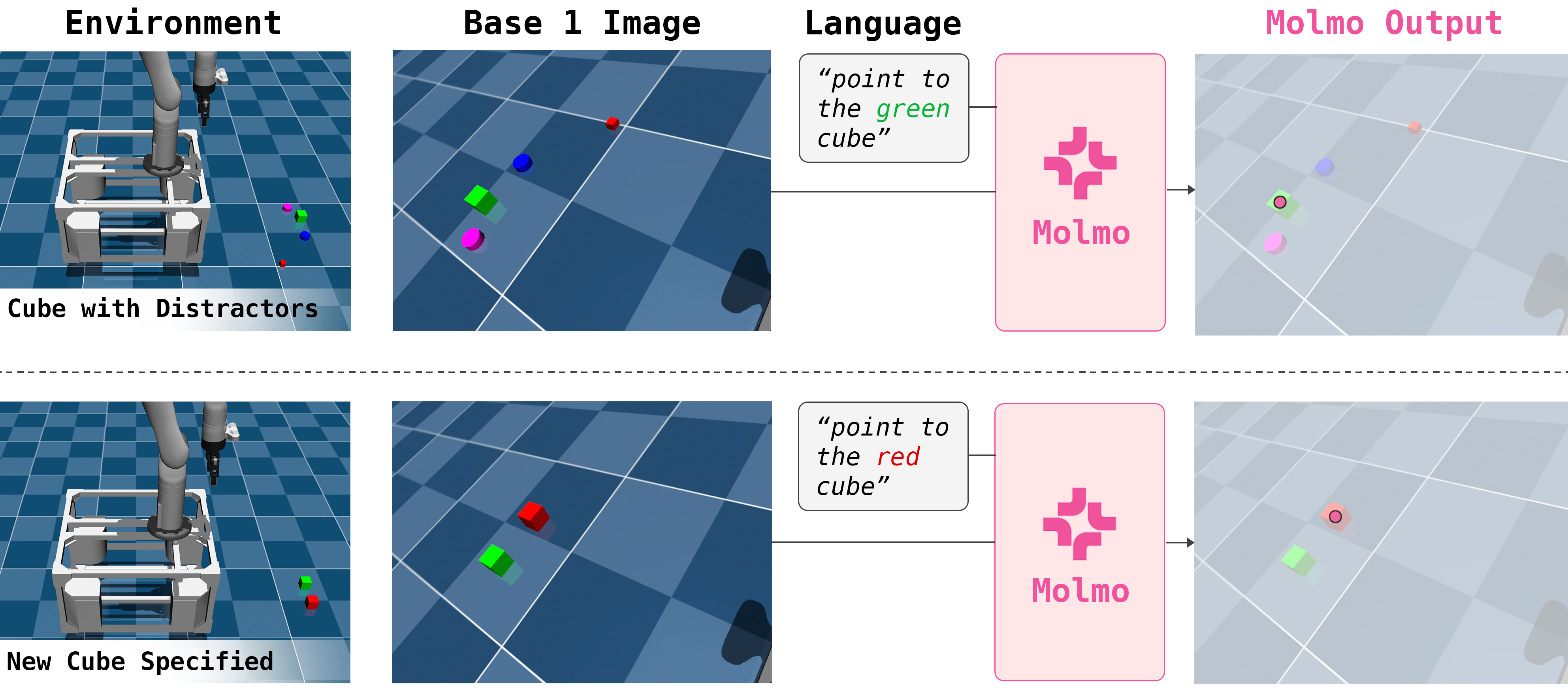}}
    \caption{\textbf{Extracting Salient Points from MolMo.} 
    In the \emph{Cube} generalization tasks (\cref{sec:q3-generalization}), we use MolMo 7B-D~\cite{deitke2024molmo} to detect task-relevant keypoints (\textcolor{red!25!pink}{\ding{108}}) from third-person images given language prompts like “point to the green cube.” The predicted pixel is backprojected into the 3D point cloud and used as the salient point input to the keypose policy. The top row shows correct selection among distractors; the bottom row shows generalization to a novel red cube.}
    \label{fig:molmo_output}
\end{figure*}
\vspace{0.3cm}

In our experiments (\cref{sec:q3-generalization}), we consider variants of the \emph{Cube} task, where the goal is to pick up a cube subject to different environment variations. Salient points are extracted using MolMo 7B-D~\cite{deitke2024molmo}, a vision-language model that returns pixel-level keypoints given a language prompt. We use the prompt “point to the $<$green/red$>$ cube,” and backproject the returned pixel into 3D to obtain the salient point (\cref{fig:molmo_output}).

While we use a fixed prompt for this simple, single-object task, our architecture is agnostic to the exact number of salient points and which objects or object parts they refer to. Future work can explore more complex settings that involve multiple objects, dynamic keypoint selection, and more general VLM prompting strategies that evolve with the task phase.

\subsection{Dense Policy}
\label{sec:appendix-dense}

We implement the dense policy of \acro~using a diffusion model that predicts fine-grained delta end-effector motions. Following~\cite{chi2023diffusion}, we use a ResNet-18 encoder to process RGB images and append proprioceptive features before passing them to a 1D convolutional UNet denoiser. The model is trained using DDPM to predict noise added to delta action sequences. 

At test-time, the policy predicts a future horizon of 16 actions and executes the first 8 before replanning. Observations include third-person and wrist-mounted RGB images. We train the model using the Adam optimizer with cosine learning rate decay and weight decay regularization. The policy used for evaluation is the final saved checkpoint.

\end{document}